\documentclass[lettersize,journal]{IEEEtran}
\usepackage{amsmath,amsfonts}
\usepackage{algorithmic}
\usepackage{algorithm}
\usepackage{array}
\usepackage[caption=false,font=normalsize,labelfont=sf,textfont=sf]{subfig}
\usepackage{textcomp}
\usepackage{stfloats}
\usepackage{url}
\usepackage{verbatim}
\usepackage{graphicx}
\usepackage{cite}
\usepackage{url}            
\usepackage{booktabs} 
\usepackage{amsfonts}       
\usepackage{mathtools}
\usepackage{nicefrac}
\usepackage{amsthm}
\usepackage{tcolorbox}
\tcbuselibrary{breakable}
\usepackage[shortlabels]{enumitem}
\usepackage[symbol]{footmisc}
\usepackage{pifont}
\usepackage{wrapfig}
\usepackage{tablefootnote}
\usepackage{makecell}
\usepackage[table,xcdraw]{colortbl}

\def\div{{\mathrm{div}}}
\newtheorem{lem}{Lemma}
\newtheorem{prop}{Proposition}

\newtheorem{defn}{Definition}

\usepackage{hyperref}
\usepackage{xcolor}
\usepackage[normalem]{ulem}
\newcounter{gaocomm}

\definecolor{blue-violet}{rgb}{0.00,0.75,0.90}
\definecolor{mygreen}{rgb}{0.0, 0.5, 0.0}
\definecolor{awesome}{rgb}{1.0, 0.13, 0.32}
\definecolor{bostonuniversityred}{rgb}{0.8, 0.0, 0.0}

\renewcommand\arraystretch{1.4}

\definecolor{LightGray}{gray}{0.9}

\newtcbox{\codebox}{on line,
  colframe=gray!80,
  colback=white,
  boxrule=0.5pt,
  arc=3pt, 
  boxsep=1pt,
  left=2pt,
  right=2pt,
  top=1pt,
  bottom=1pt,
  fontupper=\ttfamily
}

\definecolor{gre}{RGB}{193, 225, 193}
\definecolor{gre2}{RGB}{0, 128, 0}

\newtcolorbox{myprop}{
  colback=green!10,
  enhanced,
  title=General Comparison Between Methods,
  attach boxed title to top left,
  fonttitle=\bfseries,
  coltitle=black,
  toprule=0pt,
  bottomrule=0pt,
  rightrule=0pt,
  leftrule=3pt,
  arc=0mm,
  skin=enhancedlast jigsaw,
  sharp corners,
  colframe=gre,
  colbacktitle=gre,
  boxed title style={
    frame code={
      \fill[gre] 
        (frame.south west) -- 
        (frame.north west) -- 
        (frame.north east) -- 
        ([xshift=3mm]frame.east) -- 
        (frame.south east) -- 
        cycle;

      \draw[line width=1mm, gre] 
        ([xshift=2mm]frame.north east) -- 
        ([xshift=5mm]frame.east) -- 
        ([xshift=2mm]frame.south east);

      \draw[line width=1mm, gre] 
        ([xshift=5mm]frame.north east) -- 
        ([xshift=8mm]frame.east) -- 
        ([xshift=5mm]frame.south east);

    }
  }
}

\newtcolorbox{mytheorem}{
  colback=green!10,
  enhanced,
  title=Theorem 5.2 Probabilistic Upper Bound of the Parameter Gap,
  attach boxed title to top left,
  fonttitle=\bfseries,
  coltitle=black,
  toprule=0pt,
  bottomrule=0pt,
  rightrule=0pt,
  leftrule=3pt,
  arc=0mm,
  skin=enhancedlast jigsaw,
  sharp corners,
  colframe=gre,
  colbacktitle=gre,
  boxed title style={
    frame code={
      \fill[gre] 
        (frame.south west) -- 
        (frame.north west) -- 
        (frame.north east) -- 
        ([xshift=3mm]frame.east) -- 
        (frame.south east) -- 
        cycle;

      \draw[line width=1mm, gre] 
        ([xshift=2mm]frame.north east) -- 
        ([xshift=5mm]frame.east) -- 
        ([xshift=2mm]frame.south east);

      \draw[line width=1mm, gre] 
        ([xshift=5mm]frame.north east) -- 
        ([xshift=8mm]frame.east) -- 
        ([xshift=5mm]frame.south east);

    }
  }
}

\newtcolorbox{custombox}[1]{%
  enhanced,
  breakable,
  colback=green!10,        
  colframe=green!50!black, 
  coltitle=black,
  title=#1,
  fonttitle=\bfseries,
  colbacktitle=gre,   
  boxrule=0.8pt,
  arc=2mm,
  top=1mm,
  bottom=1mm,
  left=2mm,
  right=2mm
}

\newcommand{\revision}[1]{{\color{black}#1}}

\begin{document}

\title{Exposition on Over-squashing Problem of GNNs: Current Methods, Benchmarks and Challenges}
\author{Dai Shi, Andi Han, Lequan Lin, Yi Guo, Junbin Gao
\IEEEcompsocitemizethanks{
\IEEEcompsocthanksitem  
        \indent Dai Shi, Andi Han, Lequan Lin, and Junbin Gao are from the University of Sydney. Corresponding to:  \texttt{dai.shi@sydney.edu.au}. \\
        Yi Guo is with School of Computer, Data and Mathematical Sciences, Western Sydney University, Parramatta, NSW 2150, Australia. E-mail:  \texttt{y.guo@western.sydney.edu.au}.\\
        We sincerely thank Luke Thompson for his support on fixing grammar and maintaining paper consistency during the revision period.
}}


\markboth{IEEE TRANSACTIONS ON PATTERN ANALYSIS AND MACHINE INTELLIGENCE~Vol.~XX, No.~X, Aug~2026}%
{Dai \MakeLowercase{\textit{et al.}}: }


\maketitle

\begin{abstract}
Graph-based message-passing neural networks (MPNNs) have achieved remarkable success in both node and graph-level tasks. However, several identified problems, including over-smoothing (OSM), limited expressive power, and over-squashing (OSQ), still restrain the performance of MPNNs. 
In particular, the latest identified problem, OSQ, reveals that MPNNs generally fail to maintain their learning accuracy with tasks that require long-range dependencies between node pairs. In this work, we present an exposition of the OSQ problem by summarizing its various formulations in the current literature and categorizing existing solutions into three different types. In addition, we also discuss the alignment between OSQ and expressive power plus the trade-off between OSQ and OSM. Furthermore, we summarize the empirical methods
proposed by existing works to verify the efficiency of OSQ mitigation approaches, together with illustrations of their computational complexities. Lastly, we identify some open questions that are of interest for further exploration of the OSQ problem.
\end{abstract}

\begin{IEEEkeywords}
Graph Neural Networks, Over-squashing.
\end{IEEEkeywords}

\section{Introduction}
Graph message passing neural networks (MPNNs) have achieved remarkable success in terms of both node and graph level classification tasks \cite{wu2020comprehensive,yuan2022explainability,xie2022self}. Despite these successes, there are several major problems, such as over-smoothing (OSM) \cite{new_oono2020graph}, limited expressive power \cite{xu2018powerful}, and over-squashing (OSQ) \cite{topping2021understanding,alon2020bottleneck} that restrict their performance. Established from the earlier days, OSM and limited expressive problems have been well studied, and many solutions have been proposed to alleviate these problems \cite{rusch2023survey,xu2018powerful,new_zhang2023rethinking}. However, the OSQ problem, recently identified in \cite{topping2021understanding}, remains a mysterious and perplexing topic in the machine learning community. 
Initially discovered from empirical observations in \cite{alon2020bottleneck}, the OSQ problem can be interpreted as a phenomenon of information distortion. In deep MPNNs, the rich information from long-range nodes
becomes overly compressed into a limited information pack due to the graph connectivity and MPNN architecture \cite{topping2021understanding,liang2023long}. This leads to the fact that nodes distant from each other fail to transmit their messages appropriately, causing MPNNs to perform poorly in tasks that require long-term interactions. 

Although forming an intuition about OSQ is straightforward, quantifying the OSQ problem remains a challenge at the forefront of studies on MPNNs. Unlike the OSM problem, one may refer to the so-called Dirichlet energy or its variants to explicitly measure the total variation of the node features \cite{han2022generalized,digiovanni2023understanding,cai2020note}, and identify the problem if an exponential energy decay appears such that all node features tend to be the same asymptotically \cite{rusch2023survey}. 
For the OSQ problem, as an example, \cite{black2023understanding} tries to conduct an analysis over the sensitivity of each node feature generated from one specific layer of MPNN with respect to the initial features of other nodes, i.e., upper-bounding the Jacobian matrix norm with graph topological indicators such as effective resistance \cite{black2023understanding} and commute time \cite{di2023over}. However, an upper-bounded quantity does not necessarily measure the ``actual'' OSQ behaviors.

On the other hand, exploring the OSQ problem is also difficult in the corresponding empirical studies. Without a concrete numerical measure of OSQ (compared to OSM), it is challenging to verify whether an MPNN truly benefits from OSQ-mitigating approaches. 
In fact, it has been found that many existing datasets, such as ZINC \cite{new_dwivedi2023benchmarking}, \texttt{ogbg-molpcba} and \texttt{ogbg-molhiv} \cite{hu2020open}, rely more heavily on local information than distant information, making them unsuitable to illustrate the power of OSQ-mitigating methods \cite{dwivedi2022long}. 
Therefore, we claim that a group of benchmark datasets that require MPNNs to capture long-range interactions is urgently needed\footnote{For example, a dataset that requires MPNN to make a correct prediction once the number of layers $\ell$ equals the graph diameter, i.e., the longest shortest path on the graph.}. 
Although the recent contribution from \cite{dwivedi2022long} provides five such graph datasets for long-range node/graph level classification/regression tasks,  
\revision{Long Range Graph Benchmark (LRGB) is not frequently leveraged to verify the model's effectiveness, even in recent works \cite{gutteridge2023drew,new_fesser2023mitigating,new_barbero2023locality,geisler2024spatio,southern2025understanding}.}


Motivated by the aforementioned limitations, it is natural to \textbf{build up a systematic framework for the OSQ problem in graph-based MPNNs, providing insights into current methods and challenges}. In this work, we begin by illustrating various formulations of the OSQ problem and their relationships, followed by a summary of the current methods for solving the OSQ problem. We identify three general types of methods, namely \textit{spatial}, \textit{spectral}, and \textit{implicit rewiring} methods. While spatial methods aim to densify the graph locally via the identified topological indicators (i.e., edge curvature) \cite{topping2021understanding}, the spectral methods focus alternatively on building a direct connection between nodes with long-range distances so that some spectral indicators (i.e., effective resistance) can be changed accordingly \cite{black2023understanding}. Implicit rewiring methods do not propagate node features through a rewired graph. \revision{Instead, the rewiring process is conducted either within full-attentive transformers \cite{ying2021transformers}, by retrospecting node features from previous layers \cite{toth2022capturing}, or through adjustments of the information flow \cite{geisler2024spatio,bause2024two,shen2024handling}.
These methods are implicitly equivalent to those that link two distant nodes together for propagation, and are therefore included in our survey. }
Despite the diverse motivations for these methods, we will establish potential connections among them.

In addition, we draw attention to the theoretical aspect of the OSQ problem, particularly its relationship with the expressive power of MPNNs and the trade-off with the OSM problem. Transitioning seamlessly from the theoretical aspect of OSQ, we delve into a capsulization of empirical strategies to validate methods targeting OSQ mitigation, including a review of synthetic benchmarks, classic node/graph classification datasets, and long-range benchmarks. We also include a summary of computational complexity, which is a common concern and an essential aspect to address in current methods. Finally, we outline the challenges and open questions from both theoretical and empirical perspectives of OSQ, encompassing our unified definition of OSQ, along with our best-informed suggestions.

\paragraph{Related Surveys on MPNN Problems}
Several attempts have been made to summarize the current understanding of MPNN OSM, OSQ, and their expressive power. For example, \cite{rusch2023survey} provided a formal definition of the OSM problem with an additional comparison between current OSM measures, such as Dirichlet energy and $p$-Dirichlet energy \cite{shao2023enhancing}. The OSM problem has been identified as a cause of unsatisfactory learning outcomes in heterophilic graphs, where connected nodes are unlikely to share the same label \cite{digiovanni2023understanding}. The work \cite{zheng2022graph} provides a further systematic review of current MPNNs for heterophilic graphs, with additional discussion on future directions. Regarding expressive capabilities, \cite{sato2020survey} initially examined the models developed in prior studies in terms of their separation and approximation capabilities. Subsequent work in \cite{new_zhang2025expressive} provided a more comprehensive analysis of model expressiveness by evaluating capabilities such as subgraph counting and spectral decomposition. 
\revision{
\cite{new_akansha2025over} summarized existing methods on the OSQ problem, leveraging benchmarks to empirically verify methods for OSQ mitigation. \cite{attali2024rewiring} focused on preprocessing-based graph rewiring techniques. Our work extends both \cite{new_akansha2025over} and \cite{attali2024rewiring} by providing a rigorous formulation and bounds of OSQ, a comprehensive overview of current methods and their connectivities, theoretical discussions linking OSQ with other MPNN problems, empirical strategies, and finally open questions with additional recommendations on the future directions. 
}  
\revision{A recent position paper \cite{arnaiz2025oversmoothing} delineates several related notions: over-squashing, long-range effects, computational bottlenecks, and topological bottlenecks. In contrast, our survey adopts a  more theoretical perspective on over-squashing by formalizing its definition, summarizing measures, and mitigating strategies, thereby complementing the position paper.}

\revision{
\paragraph{Paper Search and Selection Criteria} \textbf{(1) Search terms:} We searched {Google Scholar} with \textit{over-squashing and graph neural networks}, together with related terms such as \textit{bottleneck}, \textit{rewiring}, and \textit{graph curvature}, covering conference, journal, and preprint literature up to mid-2025. \textbf{(2) Selection Criteria:} We prioritize papers from top-tier ML conferences and journals, such as ICML, NeurIPS, ICLR, NNLS, and PAMI, and include directly relevant OSQ preprints with sufficient methodological and empirical details. \textbf{(3) Extensions:} As OSQ is relatively recent, some works address it implicitly or via equivalent methods, such as dense or learnable adjacency. We therefore reviewed recent GNN studies, including implicit rewiring methods (Section~\ref{sec:implicit_rewiring}), and expanded coverage through reference tracing.
}



    


\begin{table}[t!]
\centering
\caption{Necessary notations and terminologies}
\label{notations}
\scalebox{0.85}{
\setlength{\tabcolsep}{4.5pt} 
\renewcommand{\arraystretch}{1.5} 
\begin{tabular}{cc}
\hline
\textbf{Notations} & \textbf{Brief Interpretation} \\ \hline
\text{$\mathbf X \in \mathbb R^{N\times c_0}$}      & Initial node feature matrix               \\
\textbf{$\mathbf{H}^{(\ell)} \in \mathbb R^{N\times c_\ell}$}    & Feature representation on $\ell$-th layer of GNN model \\
\textbf{$\mathbf{h}^{(\ell)}_i \in \mathbb R^{c_\ell}$}    & Feature vector of node $i$ at $\ell$-th layer\\

\text{$\mathbf D \in \mathbb R^{N\times N}$}      & Graph degree matrix               \\
\text{$\widehat{\mathbf{A}} \in \mathbb R^{N\times N}$}      & Normalized adjacency matrix              \\
\text{}
\text{${\mathbf{A}}(\mathbf{h}_i, \mathbf{h}_j) \in \mathbb R^{N\times N}$}      & Graph adjacency re-weighted by attention coefficients\\
\text{$\widehat{\mathbf L} \in \mathbb R^{N\times N}$}      & Normalized Laplacian matrix              \\
\text{$\{ (\lambda_i, \mathbf u_i) \}_{i=1}^N$} & Set of eigenpairs of $\widehat{\mathbf L}$ \\
\text{$\widehat{\mathbf L}^+ \in \mathbb R^{N\times N}$} & Persudoinverse of $\widehat{\mathbf L}$ \\
\text{$d_\mathcal G(i,j)$} & Standard shortest path distance between nodes $i$ and $j$. \\ 
\text{$\mathcal S_{r+1}(i)$}    & Set of neighbours of node $i$ at receptive hop $r+1 \in \mathbb N$\\
\text{$\mathcal B_{r+1}(i)$}    & Receptive field of an $(r+1)$-layer MPNN at node $i$\\
\text{$\kappa(i,j)$}  & Ollivier-Ricci curvature \cite{ollivier2007ricci} \\
\text{$F(i,j)$} & Forman Ricci curvature \cite{forman2003discrete,sreejith2016forman}\\
\text{$\mathrm{BFC}(i,j)$} &Balanced Forman curvature \cite{topping2021understanding} \\
\text{$\mathrm{JLC}(i,j)$} & Jost and Liu Curvature \cite{giraldo2022understanding} \\
\text{$\textbf{AFR}_3(i,j)$,\,\,$\textbf{AFR}_4(i,j)$} &Augmented Forman curvature \cite{new_fesser2024augmentations}\\
\text{$R_{i,j}$}  &Effective resistance of between node $i$ and $j$ \cite{black2023understanding}\\
\text{$\mathrm{Com}(i,j)$} & Commute time between node $i$ and $j$ \\
\hline
\end{tabular}}
\end{table}

\paragraph{Organization}
The rest of the paper is organized as follows. Section \ref{preliminary} provides basic notations of graphs, MPNNs, the formulation, and bounds
of the OSQ problem. Section \ref{current_methods} summarizes the current methods that mitigate the OSQ problem, along with a discussion of the links between them. In Section \ref{sec:osq_theory}, we review the current theoretical understanding of the relationship between OSQ and expressive power, as well as the OSM problem, followed by Section \ref{sec:empirical_strategy}, in which most existing datasets for verifying the functionality of OSQ mitigation methods are listed. Lastly, in Section \ref{open_questions}, we outline the current challenges related to the OSQ problem, along with some suggestions based on our understanding and knowledge. {To support readability, each main technical section concludes with a concise “Takeaway Message” that highlights the central points of the discussion.}

\section{Message Passing Paradigm and Over-squashing}\label{preliminary}
In this section, we provide notations on graphs and the formulation of MPNNs. In addition, we will propose a new definition of the OSQ on the graph and demonstrate how the effect of the OSQ problem is bounded using various indicators from recent literature.   To avoid any confusion in terms of the notations and terminologies, we list them in Table \ref{notations}.

\subsection{Graph and MPNNs}
We start by introducing notations used throughout this paper. A graph is a tuple $\mathcal G = (\mathcal V, \mathcal E)$ where $\mathcal V$ and $\mathcal E$ denote the set of nodes and edges, respectively. The nodes are indexed by $i$ and $j$ such that $i, j\in\mathcal{V}$, $i,j = 1,2,\ldots, N = |\mathcal V|$, and an edge connecting nodes $i$ (from) and $j$ (to) is denoted by $(i,j)\in \mathcal{E}$. In this work, we focus on undirected graphs, i.e., if $(i,j) \in \mathcal E$, then $(j,i) \in \mathcal E$. We the define $N \times N$ adjacency matrix $\mathbf{A}$ as $\mathbf{A}_{i,j} =1$ if $(i,j) \in \mathcal E$ and zero otherwise. In addition, we let $\mathbf D$ be the diagonal matrix with $D_{ii} = d_i$, the degree of node $i$, and denote $d_{\mathrm{max}}$ and $d_{\mathrm{min}}$ as the maximal and minimal degrees. The normalized adjacency matrix is denoted by $\widehat{\mathbf{A}} = {\mathbf D}^{-1/2}(\mathbf{A} + \mathbf{I}){\mathbf D}^{-1/2}$ where node self-loops are added. 
Accordingly, the normalized graph Laplacian is given by $\widehat{\mathbf L} = \mathbf{I} - \widehat{\mathbf{A}}$. It is well known that the $\widehat{\mathbf L}$ admits the eigendecomposition $\widehat{\mathbf L} = \mathbf U\boldsymbol{\Lambda} \mathbf U^\top$, where $\mathbf U^\top \mathbf{h}$ is known as the Fourier transform of a graph signal $\mathbf{h} \in \mathbb R^N$.
In this paper, we let $\{ (\lambda_i, \mathbf u_i) \}_{i=1}^N$ be the set of eigenvalue and eigenvector pairs of $\widehat{\mathbf L}$ where $\mathbf u_i$ is the $i$th column vector of $\mathbf U$. Furthermore, we order the eigenvalues of $\widehat{\mathbf L}$ in increasing order. In $0 = \lambda_1 \leq \lambda_2 \leq \cdots \lambda_N \leq 2$, $\lambda_2$ is the so-called \textit{spectral gap} of graph $\mathcal G$, and we use the Moore-Penrose pseudoinverse of $\widehat{\mathbf L}$, defined as $\widehat{\mathbf L}^+ = \sum_{\{i:\lambda_i>0\}}\frac1\lambda_i\mathbf u_i\mathbf u_i^\top$.

\paragraph{Message passing neural networks (MPNNs)}
Assume $\mathcal G$ is equipped with a node feature matrix $\mathbf X \in \mathbb R^{N \times c_0}$, where $\mathbf x_i \in \mathbb R^{c_0}$ represents the feature vector at node $i$. We define $\mathbf{h}_i^{(\ell)}$ as the feature representation of node $i$ at layer $\ell$, with $\mathbf{h}^{(0)}_i = \mathbf x_i$. The Message Passing Neural Networks (MPNNs) \cite{gilmer2017neural} is based on a family of message mixing functions $\psi_\ell : \mathbb R^{c_\ell} \times \mathbb R^{c_\ell} \rightarrow \mathbb R^{c_\ell'}$ and update functions $\phi_\ell: \mathbb R^{c_\ell} \times \mathbb R^{c_\ell'} \rightarrow \mathbb R^{c_{\ell +1}}$ as follows:
\begin{align}\label{mpnn}
\mathbf{h}_i^{(\ell+1)} = \phi_\ell  (\mathbf{h}_i^{(\ell)}, \sum_{j \in \mathcal N_i} \widehat{\mathbf{A}}_{ij}\, \psi_\ell(\mathbf{h}_i^{(\ell)}, \mathbf{h}_j^{(\ell)})),
\end{align}
where $\mathcal N_i\subseteq \mathcal V$ is the subset containing all neighboring nodes of node $i$, a.k.a. the one-hop neighbour set, written as $\mathcal{S}_1(v_i)$. A common implementation of the functions $\phi$ and $\psi$ involves the use of activation and attention mechanisms \cite{velivckovic2018graph} to generate feature similarities. It is worth noting that recent research efforts aimed to extend traditional MPNNs in Eq.~\eqref{mpnn} through various approaches. These include message passing over a re-weighted fully connected graph \cite{new_wu2023difformer}, as well as rewired graphs with improved topological properties \cite{topping2021understanding, giraldo2022understanding}.

\subsection{The OSQ problem and OSQ Bounds}\label{OSQ_bounds_1}
\paragraph{The Over-squashing Problem (OSQ)}
The OSQ problem was first identified by \cite{alon2020bottleneck} as a phenomenon, known as a \textit{bottleneck}, emerging during the aggregation of messages by graph nodes along lengthy paths. The presence of this bottleneck in a graph leads to a phenomenon wherein exponentially more information is compressed into fixed-sized vectors when the number of layers in MPNN increases, i.e., going ``deeper'' \cite{topping2021understanding}. \textbf{Consequently, graph-based MPNNs
are unable to effectively transmit messages from distant nodes and exhibit poor performance when the prediction task relies on long-range interactions.} Although OSQ was only identified empirically in \cite{alon2020bottleneck}, the latter work by \cite{topping2021understanding} first attempted to quantify OSQ phenomenon by virtual of the sensitivity of each node representation $\mathbf{h}^{(\ell)}_i$ with respect to  $\mathbf x_s$ 
denoted by $\frac{\partial \mathbf{h}_i^{(\ell)}}{\partial \mathbf x_s}$. We call the spectral norm of the Jacobian matrix (i.e., $\| \frac{\partial \mathbf{h}_i^{(\ell)}}{\partial \mathbf x_s}\|$) the \textbf{OSQ score} to distinguish it from the aforementioned OSQ phenomenon:
\begin{defn}[OSQ Score]
The OSQ score, ${\rm OSQ}^{(\ell)}(i,s)$
between node $i, s$ 
after $\ell$ layers of message passing in \revision{Eq.~\eqref{mpnn}} is defined by, 
\revision{
\begin{align}
  {\rm OSQ}^{(\ell)}(i,s) = \left\| \frac{\partial \mathbf{h}_i^{(\ell)}}{\partial \mathbf x_s} \right \|, 
\end{align}
}
where $\|\cdot\|$ is the spectral norm of a matrix. 
\end{defn}
Furthermore, \cite{topping2021understanding} bounded the OSQ score with the following. 
\begin{prop}[OSQ Score Bound \cite{topping2021understanding}]\label{OSQ_topping}
    Assuming an MPNN denoted in Eq.~\eqref{mpnn}. Let $i,s \in \mathcal V$ with $s\in \mathcal S_{r+1}(i)$. If $\|\nabla \phi_\ell\| \leq \alpha$ and $\|\nabla \psi_\ell\| \leq \beta$ 
    for $0 \leq \ell \leq r$, then 
    \begin{align}\label{osq_problem}
        \left \|\frac{\partial \mathbf{h}_i^{(r+1)}}{\partial \mathbf x_s}\right \| \leq (\alpha\beta)^{r+1}(\widehat{\mathbf{A}}^{r+1})_{is},
    \end{align}
    where $\mathcal S_{r+1}(i)\coloneqq\{j\in \mathcal V : d_\mathcal G(i,j) = r+1\}$  is the set of neighbours of node $i$ whose standard shortest path distance to node $i$ on graph $\mathcal G$, written as $d_\mathcal G(i,j)$, is $r+1 \in \mathbb N$. Note that $\widehat{\mathbf{A}}^{r+1}$ is $\widehat{\mathbf{A}}$ to the power of $r+1$.
\end{prop}
\noindent In addition to the above definition, we also let $\mathcal B_{r+1}(i) \coloneqq \{ j \in \mathcal V: d_\mathcal G (i,j) \leq r+1\}$ be the receptive field of node $i$. The definition of the OSQ score shows that if $\nabla \phi_\ell$ and $\nabla \psi_\ell$ are bounded, then the propagation of the messages is controlled by 
of $\widehat{\mathbf{A}}$. We note that this definition is also suitable for the graph with a re-weighted adjacency matrix and a similar conclusion has been derived in \cite{new_shi2024new}. One important observation of the definition and the upper bound presented is that two nodes $i$ and $s$ are required to be \textbf{exactly} distanced with $r+1$ hops. However, in \cite{black2023understanding}, OSQ is defined in terms of any node pairs as follows,  \revision{which has been adopted in works such as \cite{saber2025over}}.
\begin{prop}[OSQ Score Bound \cite{black2023understanding}]
    With the assumptions provided in Proposition \ref{OSQ_topping}, one can bound the OSQ score between node $i$ and $s$ as 
    \begin{align}\label{osq_problem_2}
         \left \|\frac{\partial \mathbf{h}_i^{(r+1)}}{\partial \mathbf x_s}\right \| \leq (\alpha\beta)^{r+1} \sum_{\ell=0}^{r+1}
         (\widehat{\mathbf{A}}^{\ell})_{is}.
    \end{align}
\end{prop}



Another reason for bounding the OSQ score with Eq.~\eqref{osq_problem_2} is to relate it to other graph topological indicators, such as effective resistance and commute time \cite{black2023understanding,di2023over}, which are defined for any node pair. Taking its name from its use in electrical engineering contexts, the effective resistance between node $i,j$ is defined as:
\begin{equation}\label{er_formular}
\resizebox{0.4\textwidth}{!}{$
\begin{aligned}
    R_{i,j} \coloneqq 
    \left(\frac{1}{\sqrt{d_i}}\boldsymbol{1}_i 
    - \frac{1}{\sqrt{d_j}}\boldsymbol{1}_j\right)^\top 
    \widehat{\mathbf L}^+  
    \left(\frac{1}{\sqrt{d_i}}\boldsymbol{1}_i 
    - \frac{1}{\sqrt{d_j}}\boldsymbol{1}_j\right)
\end{aligned}
$}
\end{equation}
in which $\widehat{\mathbf L}^+$ is the pseudoinverse of $\widehat{\mathbf L}$, and both $\boldsymbol{1}_i$ and $\boldsymbol{1}_j$ are the indicator vectors of node $i$ and $j$, respectively. Intuitively, the effective resistance is a measure of how ``well-connected'' two nodes $i$ and $j$ are: If there are many possible paths between them, then $R_{i,j}$ tends to be small, whereas if there are few paths between them, $R_{i,j}$ tends to be larger. Thus, $R_{i,j}$ is a good candidate for describing OSQ. In fact, in \cite{black2023understanding}, the bound of OSQ score 
is further presented via the $R_{i,s}$\footnote{Different from Eq.~\eqref{er_formular}, we change the node pair from $i,j$ to $i,s$ to match the definition of OSQ provided in Eq.~\eqref{osq_problem_2}.} that is 
\begin{prop}[OSQ Score Bound \cite{black2023understanding}]
\begin{equation}\label{rr_bound}
\resizebox{0.49\textwidth}{!}{$
\begin{aligned}
     &\left \|\frac{\partial \mathbf{h}_i^{(r+1)}}{\partial \mathbf x_s}\right \| \notag 
     \leq (\alpha\beta)^{r+1} \frac{d_\mathrm {max}}{2}
     \left(\frac{2}{d_\mathrm {min}} \left(r+2 - \frac{\mu^{r+2}}{1-\mu}\right)-R_{i,s}\right),
\end{aligned}
$}
\end{equation}
where we $d_\mathrm {min}$ and $d_\mathrm {max}$ are $\mathrm{min}\{d_i, d_s\}$ and $\mathrm{max}\{d_i, d_s \}$, respectively and $\mathrm{max}\{|\mu_2|, |\mu_N|\} \leq \mu$, with $\mu_N \leq \mu_{N-1}, \cdots, < \mu_1 = 1$ are the sorted eigenvalues of $\widehat{\mathbf{A}}$.
\end{prop}
For any node pair $i$ and $s$, if they are connected with multiple paths, $R_{i,s}$ tends to be small; therefore, a higher upper bound is achieved compared to the node pairs connected with fewer paths, so that a larger resistance is induced. In fact, in \cite{black2023understanding}, under the assumption $\mathcal G$ is non-bipartite, then $R_{i,s}$ may be computed as
\begin{align}\label{rr_computation}
    R_{i,s} = \sum_{k = 0}^\infty \left(\frac{1}{d_i}(\widehat{\mathbf{A}}^k)_{ii} + \frac{1}{d_s}(\widehat{\mathbf{A}}^k)_{ss}- \frac{2}{\sqrt{d_id_s}}(\widehat{\mathbf{A}}^k)_{is}\right),
\end{align}
which is leveraged in the proof for Eq.~\eqref{rr_bound}. Compared to the upper bounds presented in Eq.~\eqref{osq_problem} and \eqref{osq_problem_2}, which bound the OSQ score 
via local edge weights (i.e.,$(\widehat{\mathbf{A}})_{is}$), the bound showed in Eq.~\eqref{rr_bound} views OSQ from a global perspective as effective resistance can be defined via any pair of nodes regardless of the number of hops between them. 

In parallel to the work of \cite{black2023understanding}, the OSQ problem is also studied via commute time, denoted as $\mathrm{Com}(i,s)$, 
which is the expected number of steps in a random walk from $i$ to $s$ and back to $i$. It is well known that the commute time is proportional to the effective resistance, that is $\mathrm{Com}(i,s) = 2|\mathcal E| R_{i,s}$, where we recall that $\mathcal E$ is the set of edges of the graph. 
In \cite{di2023over}, the OSQ score
is enhanced to the symmetric Jacobian obstruction between node $i$ and $s$. Specifically, for any layer number $k \leq \ell$, one can define
\begin{align}\label{eq:jacobian_obs}
    \widehat{\mathbf O}(\ell)_{i,s} \coloneqq \sum^{(\ell)}_{k = 0} \| (\mathbf J_k^{(\ell)})_{i,s} \|,
\end{align}
where

\resizebox{0.49\textwidth}{!}{$
\begin{aligned}
    (\mathbf J_k^{(\ell)})_{i,s}  =& \left(\frac{1}{d_i} \frac{\partial \mathbf{h}^{(\ell)}_i}{\partial \mathbf{h}^{(k)}_i} - \frac{1}{\sqrt{d_i d_s}}\frac{\partial \mathbf{h}^{(\ell)}_i}{\partial \mathbf{h}^{(k)}_s}\right) \!
    +\! \left(\frac{1}{d_s} \frac{\partial \mathbf{h}^{(\ell)}_s}{\partial \mathbf{h}^{(k)}_s} - \frac{1}{\sqrt{d_i d_s}}\frac{\partial \mathbf{h}^{(\ell)}_s}{\partial \mathbf{h}^{(k)}_i}\right).
\end{aligned}
$}
One can intuitively understand $(\mathbf J_k^{(\ell)})_{i,s}$ in a similar manner as effective resistance. For example, $\| (\mathbf J_k^{(\ell)})_{i,s}\|$ tends to be larger if nodes $i$ and $s$ are less sensitive to each other and smaller whenever the communication is sufficiently robust.
Apparently, $  \widehat{\mathbf O}(\ell)_{i,s}$ is an ``upgraded'' version of OSQ score. It is bounded by the commute time as follows.
\begin{prop}[OSQ Score Bound \cite{di2023over}] 
Let $\psi_\ell \coloneqq \mathbf W_\ell$
be the weight matrix of one specific MPNN induced from Eq.~\eqref{mpnn}, $\nu$ the minimal singular value of $\mathbf W_\ell$ and $\mathrm{w}$ the maximal spectral norm of $\mathbf W_\ell$ (i.e., $\|\mathbf W_\ell\| \leq \mathrm w$). Assume all paths in the graph of the MPNN defined in Eq.~\eqref{mpnn} are activated with the same probability of success $\rho$. \revision{Then for Jacobian obstruction defined in Eq.~\eqref{eq:jacobian_obs}} there exists a constant $\epsilon_\mathcal G$ independent of $i,s$, such that 
\begin{align}\label{our_commute_bound}
    \epsilon_\mathcal G\left(1-o(\ell)\right) \frac{\rho}{\nu} \frac{\mathrm{Com}({i,s})}{2|\mathcal E|}
    \leq \widehat{\mathbf O}(\ell)_{i,s} \leq \frac{\rho}{\mathrm{w}} \frac{\mathrm{Com}({i,s})}{2|\mathcal E|},
\end{align}
with $o(\ell) \rightarrow 0$\footnote{For the explicit form of $o(\ell)$, please see \cite{di2023over} Appendix for more details.} exponentially fast with the increase of $\ell$\footnote{We highlight that the conclusion we showed is different to the one derived in \cite{di2023over}, due to different MPNN formats.}.
\end{prop} 
Plugging in the relationship between $\mathrm{Com}$ and $R$, the above equation becomes $ \epsilon_\mathcal G\left(1-o(\ell)\right) \frac{\rho}{\nu} R_{i,s}
\leq \widehat{\mathbf O}(\ell)_{i,s} \leq \frac{\rho}{\mathrm{w}} R_{i,s}$, which shows a  similar result reported in \cite{black2023understanding}. Lastly, given the relationship between commute time and local Cheeger constant\footnote{For one graph we denote Cheeger constant as $\mathcal H (\mathcal G) = \min_{\mathcal{I} \in \mathcal V} \frac{|\{(i,s)\in \mathcal E: i \in \mathcal I, s \in \mathcal V \setminus \mathcal I\}|}{\min(\mathrm{vol(\mathcal I),\mathrm{vol(\mathcal V\setminus \mathcal I))}}}$, where $\mathrm{vol(\mathcal I} = \sum_{i\in \mathcal I} d_i)$.}, in \cite{di2023over}, the upper bound of $\widehat{\mathbf O}^{(\ell)}_{i,s}$ is further represented via the reciprocal of Cheeger constant to interpret how the rewiring methods are efficient for mitigating the OSQ problem. 


Finally, it is worth noting that the notion of OSQ can be extended to the task of graph-level classification, and in \cite{new_di2024does}, the OSQ for graph level classification task is defined via the reciprocal of maximal mixing power of one MPNN, that is 

\begin{defn}[Graph Level OSQ \cite{new_di2024does}]\label{glosq}
For a twice differentiable graph function $\mathbf Y_\mathcal G$, serving as the ground-truth function to be learned by MPNN defined in Eq.~\eqref{mpnn}. The graph level OSQ is defined as the reciprocal of maximal mixing induced by $\mathbf Y_\mathcal G$ among the features $\mathbf x_i$, $\mathbf x_s$. That is
\begin{align}\label{eq:glosq}
    \mathrm{GLOSQ}(i,s):& = \left(\mathrm{mix}_{\mathbf Y_\mathcal G}(i,s)\right)^{-1} \notag \\
    &= \left(\max_{\mathbf x_i} \max_{1 \leq \gamma, \delta \leq c_0}\left|\frac{\partial^2 (\mathbf Y_\mathcal G(\mathbf X))}{\partial\mathbf x_i[\gamma] \partial \mathbf x_s[\delta]}\right| \,\right)^{-1},
\end{align}
where $\mathbf x_i[\gamma]$ and $\mathbf x_s[\delta]$ are the $\gamma$-th and $\delta$-th component of the feature of node $i$ and $s$, respectively\footnote{It is worth noting that the notation $(i,s)$ here means the node $i,s$ pair rather than the edge connected with node $i$ and $s$.}. 
\end{defn}
\revision{One can check the mixing power (i.e., $\mathrm{mix}_{\mathbf Y_\mathcal G}$) in Eq.~\eqref{eq:glosq} measures the ability of a model to mix features associated with different nodes, showing the model's \textit{expressive power} as described in \cite{new_di2024does}. Intuitively, it quantifies the strongest pairwise coupling the task can induce: larger maximal mixing implies easier long-range interaction, whereas smaller maximal mixing corresponds to stronger over-squashing.
} We further discuss the relation between maximal mixing power and OSQ in Section \ref{sec:osq_theory}. It is worth noting that there are several additional measures for OSQ, such as the influential factor defined in \cite{xu2018representation}, the squashing coefficient in \cite{sun2022position}, the traditional bottleneck value \cite{topping2021understanding}, \revision{and node distance in \cite{li2024addressing}.} \revision{In this paper, our main focus is on the effect of OSQ in node-level classification tasks on static graphs, although recently OSQ problem has also been discussed in hypergraphs \cite{new_yadati2024oversquashing}.} 

\begin{tcolorbox}[breakable]
\textbf{Takeaway Message of This Section:} \textit{OSQ refers to the phenomenon where exponentially increasing feature information is compressed into fixed-sized vectors due to the graph's connectivity. The OSQ score, denoted as $ {\rm OSQ}^{(\ell)}(i,s) = \left\| \frac{\partial \mathbf{h}_i^{(\ell)}}{\partial \mathbf x_s} \right \|$, can be upper bounded by topological indicators such as graph curvature, effective resistance and commute time.}
\end{tcolorbox}



\section{Strategies that Mitigates OSQ}\label{current_methods}
\revision{
In this section, we review the strategies that mitigate the OSQ problem. In general, these methods can be categorized into three categories: (1)  \textit{Spatial rewiring} directly focuses on the graph spatial connectivity, i.e., adjacency, usually builds edges between nodes in their $r$-hop radius \cite{topping2021understanding,giraldo2022understanding}. (2) \textit{Spectral rewiring} builds edges between nodes so that some graph global expansion properties, such as spectral gap \cite{new_karhadkar2023fosr}, total effective resistance \cite{black2023understanding} and commute time \cite{new_di2024does}, are optimized. (3) \textit{Implicit rewiring}, for which better graph connectivities (e.g., denser, reweighted, full-connected graph) are constructed to benefit purposes other than the OSQ problem, e.g., graph transformer and its variants \cite{ying2021transformers,wu2022nodeformer,new_wu2023difformer}, diffusion-based MPNNs \cite{thorpe2022grand,chamberlain2021beltrami,toth2022capturing}, and multi-track feature propagation methods \cite{pei2024multitrack,new_choi2024panda}. 
    


It is worth noting that, even though the OSQ problem might not be the primary focus in some of those papers, especially for those implicit rewiring methods, we have included them with the principle that, in general, graphs with denser connectivities are more robust to the OSQ problem compared to their sparse counterparts \cite{dwivedi2022long}. 
}


\subsection{Spatial Rewiring Techniques}\label{sec:spatial_rewiring}

\revision{

After \cite{alon2020bottleneck} empirically observed the OSQ problem, the natural next question is how to identify a reliable indicator of OSQ. \cite{topping2021understanding} address this question by initially connecting OSQ with graph Ricci curvature, showing that edges with highly negative Ricci curvature are principal contributors to OSQ. 

In differential geometry, curvature describes whether nearby geodesics tend to converge or diverge; positive Ricci curvature indicates local contraction and negative curvature indicates expansion. On graphs, discrete curvature plays an analogous role by assigning to each edge \((i,j)\) a value that reflects how similar or redundant the neighborhoods of \(i\) and \(j\) are. Highly negative curvature flags edges that act as narrow bridges between otherwise weakly connected regions, along which messages must traverse long effective distances and thus induce OSQ. In contrast, highly positive curvature corresponds to dense and redundant local connectivity, where information mixes quickly, which can relate to OSM.

In general, two principal discrete analogues of Ricci curvature are used in graphs. \textit{Forman curvature} \(F(i,j)\) \cite{forman2003discrete} is combinatorial and depends on local degrees and small motifs such as shared triangles \cite{sreejith2016forman}. \textit{Ollivier–Ricci curvature} \(\kappa(i,j)\) \cite{ollivier2007ricci} is transport-based and compares one-step neighborhood distributions via optimal transport \cite{bauer2011ollivier,lin2011ricci}. While \(F(i,j)\) is simple to compute but can be biased toward negative values, \(\kappa(i,j)\) is conceptually close to classical Ricci curvature yet more expensive and harder to localize \cite{topping2021understanding}. Given the central role of curvature as an indicator for OSQ, several adjusted formulations have been proposed \cite{topping2021understanding,shi2023curvature,bauer2011ollivier,lin2011ricci,giraldo2022understanding}. For completeness, we reproduce below the original definitions of Forman curvature \cite{sreejith2016forman} and Ollivier–Ricci curvature \cite{ollivier2007ricci}.

}

\begin{defn}[Ollivier-Ricci Curvature \cite{ollivier2007ricci}]\label{def:ollivierriccicurvature}
    Let $m$ be a family of probability measures on node $j \in \mathcal V$ depending measurably on
node $i$ with finite first momentum. For example, for $\alpha \in [0,1]$, we can have
    \begin{equation*}
    m_i(j) = 
    \begin{cases}
        \alpha, & j = i \\
        \frac{1 - \alpha}{|\mathcal{N}_i|}, & j \in \mathcal{N}_i \\
        0, & \text{ otherwise}
    \end{cases}
\end{equation*}
     Then the graph Ollivier-Ricci curvature between nodes $i$ and $j$ is defined as
\begin{equation}\label{ollivier_original}
    \kappa(i,j) = 1 - \frac{W_1(i,j)}{d_\mathcal G(i,j)},
\end{equation}
where $d_\mathcal G(i,j)$ is the shortest path distance on $\mathcal{G}$ between nodes $i$ and $j$ and $W_1(i,j)$ is the $L_1$-Wasserstein distance defined as $W_1(i,j) = \inf_{\Gamma} \sum_{i'} \sum_{j'} \Gamma_{i'j'} d_\mathcal G(i',j')$ where $\Gamma$ is the joint distribution satisfying the coupling conditions, i.e., $\sum_{i'}\Gamma_{i'j'} = m_j(j')$, $\sum_{j'} \Gamma_{i'j'} = m_i(i')$ for all $i',j'$.
\end{defn}
It is worth highlighting that there exist many alternatives to define $m_i$ as long as they satisfy the conditions in Def. \ref{def:ollivierriccicurvature}. \revision{Furthermore, based on the formulation of $\kappa$ in Eq.~\eqref{ollivier_original}, for a graph with edge number $|\mathcal E|$, the computational complexity of $\kappa$  is known to run in cubic time, i.e., $\mathcal O(d_\mathrm{max}^3|\mathcal E|)$ where $d_\mathrm{max}$ is maximal degree of vertices on the graph \cite{waissi1994network,sia2019ollivier}. In practice, one can reduce this complexity by applying the approximation algorithms such as Sinkhorn, and the curvature information is only needed to compute once before MPNN.
}


\begin{defn}[Combinatorial Forman Curvature \cite{forman2003discrete,sreejith2016forman}]
    For any edge $(i,j) \in \mathcal E$ the Forman curvature is given by
    \begin{align}\label{forman_original}
        F(i,j)\coloneqq 4- d_i - d_j + 3|\sharp_\Delta(i,j)|,
    \end{align}
    where ${\sharp}_\Delta(i,j)\coloneqq \mathcal S_1(i) \cap \mathcal S_1(j)$ 
    stands for the triangles that share edge $i\sim j$.
\end{defn} 
\revision{By counting triangles via per-edge neighborhood intersections, the complexity $F(i,j)$ for a whole graph is $\mathcal O(\sum_{v \in \mathcal V} d(v)^2)$ and it is upper bounded (the worst case) by $\mathcal O(|\mathcal E|d_{\mathrm{max}})$  \cite{new_fesser2024augmentations}.}  We further note that the definition of Forman curvature can be extended to the structures (known as CW complex) higher than triangles, i.e., quadrangles \cite{new_fesser2023mitigating}. These complexes contain rich graph structural information such that one may expect ${\sharp}_\Delta(i,j)$ to be related to positive curvature (i.e., complete graph), quadrangles as zero curvature (i.e., grid graph) and other shapes which may lead to negative curvatures. Motivated by this purpose, the balanced Forman curvature is defined in \cite{topping2021understanding} and we start by reviewing it as follows. 

\paragraph{{Stochastic Discrete Ricci Flow (SDRF)} \cite{topping2021understanding}}
As the first work that links curvature information to the OSQ problem, \cite{topping2021understanding} developed one curvature graph rewiring algorithm named as Stochastic Discrete Ricci Flow (SDRF) to mitigate the OSQ problem. One adjusted version of Forman curvature, named as \textit{balanced Forman curvature} (BFC) is defined as:
\begin{equation}
\resizebox{0.5\textwidth}{!}{$
\begin{aligned}
    \mathrm{BFC}(i,j)
    :=& \frac{2}{d_i} + \frac{2}{d_j} - 2 
    + 2\frac{|\sharp_\Delta(i,j)|}{\mathrm{max}(d_i, d_j)} 
    + \frac{|\sharp_\Delta(i,j)|}{\mathrm{min}(d_i, d_j)}
    + \frac{(\gamma_\mathrm{max})^{-1}}{\mathrm{max}(d_i, d_j)}
    \left(|\sharp^i_\Box| + |\sharp^j_\Box|\right)
\end{aligned}
$}
\end{equation}
where $\sharp^i_\Box(i,j) : = \{ k \in \mathcal S_1(i) \setminus \mathcal S_1(j), k \neq j : \exists w \in (\mathcal S_1(k) \cap \mathcal S_1(j)) \setminus \mathcal S_1(i) \}$ are the neighbors of $i$ forming a $4$-cycle based at the edge $(i,j)$ without diagonals inside, $\sharp^i_\Box$ is its short form omitting $(i,j)$, and  $\gamma_{\mathrm{max}}(i,j)$ is the maximal number of $4$-cycles based at $(i,j)$ traversing a common node, that is
\begin{align*}
    \gamma_{\mathrm{max}}(i,j): = &\mathrm{max}\left \{\mathrm{max}_{k \in \sharp^i_\Box}\{(\mathbf{A}_k \cdot (\mathbf{A}_j - \mathbf{A}_i \odot \mathbf{A}_j )) -1\}, \right. \\
    &\left.\mathrm{max}_{w \in \sharp^j_\Box}\{(\mathbf{A}_w \cdot (\mathbf{A}_i - \mathbf{A}_j \odot \mathbf{A}_i )) -1\} \right\},
\end{align*}
where $\mathbf{A}_s$ denotes the $s$-row of adjacency matrix, \revision{and as $\mathrm{BFC}$ consider the 4-cycles and $\gamma_{\mathrm{max}}$ additional to the original formulation in Eq.~\eqref{forman_original}, the computational complexity (in the worst case) for BFC becomes $\mathcal O(|\mathcal E|d_{\mathrm{max}}^2)$.
} Based on the definition of BFC, \cite{topping2021understanding}  proved that it is those edges with highly negative BFC that are responsible for the OSQ problem, and SDRF is conducted by adding edges to ``support'' the graph's most negatively curved edge and then removing the most positively curved edge. It is worth noting that the whole SDRF is merely a pre-processing method, resulting in a rewired graph that is more robust to the OSQ problem.
In addition, several important observations from BFC and SDRF are: 
\begin{enumerate}
    \item Once the graph becomes denser from SDRF, the quantity of those highly negative BFC decreases, and the OSQ problem is reduced and the message passing between nodes becomes more fluent. However, this may lead to the so-called over-smoothing (OSM) \cite{new_oono2020graph} problem, since the node features will be homogenized from MPNNs with more convenience paths. This observation suggests a potential trade-off between OSM and OSQ.
    \item  As a prepossessing method, SDRF only generates the suitable graph for the initial stage of MPNN, it is unknown whether the graph structure from SDRF is still efficient during the training of MPNN. 

    \item In order to conduct SDRF, or in general, to obtain the desired edge curvature on the modified graph, a large computational cost is expected. It is natural to consider a cheaper method that achieves the same goal as SDRF.
\end{enumerate}

Different from the OSQ problem, the OSM problem in GNN has been identified and attracted much richer research \cite{wu2020comprehensive,han2022generalized,digiovanni2023understanding}. Since adding edges alleviates the OSQ problem at the cost of exacerbating the OSM problem, it is natural to ask: \textit{ Can curvature also serve as an indicator of OSM problem? If so, is there a way of using curvature to mitigate both OSM and OSQ in a more controllable manner? }
\paragraph{Stochastic Jost and Liu Curvature Rewiring (SJLR) \cite{giraldo2022understanding}}
Building upon the above research question, \cite{giraldo2022understanding} discussed the \textit{trade-off} between the OSM and OSQ problem through the \textit{Jost and Liu Curvature} defined as
\begin{equation}
\resizebox{0.5\textwidth}{!}{$
\begin{aligned}
    \mathrm{JLC}(i,j)
    :=&-\left(1-\frac{1}{d_i} - \frac{1}{d_j} - \frac{|\sharp_\Delta(i,j)|}{d_i \wedge d_j} \right)_+ 
    - \left(1-\frac{1}{d_i} - \frac{1}{d_j} - \frac{|\sharp_\Delta(i,j)|}{d_i \vee d_j} \right)_+ 
    + \frac{|\sharp_\Delta(i,j)|}{d_i \vee d_j}
\end{aligned}
$}
\end{equation}
where $c_+ \coloneqq \mathrm{max}(c,0)$, $c \vee t \coloneqq \mathrm{max}(c,t)$ and $c \wedge t : = \mathrm{min}(c,t)$. Similar to \cite{topping2021understanding}, 
\revision{
a JLC-based graph-rewiring method called Stochastic Jost–Liu Curvature Rewiring (SJLR) was proposed in \cite{giraldo2022understanding}; it dynamically adds edges near regions of highly negative curvature to alleviate OSQ and removes edges exhibiting highly positive curvature to mitigate OSM. Finally, one can see the JLC maintains the complexity of Forman curvature (i.e., $\mathcal O(\sum_{v \in \mathcal V} d(v)^2)$).
}

Furthermore, it is worth noting that different from SDRF which serves as a pre-processing method, SJLR dynamically adds/removes edges during training so that it can alleviate OSM and OSQ without modifying the initial graph.

\paragraph{Batch Ollivier-Ricci Flow (BORF) \cite{nguyen2023revisiting}}
As both SDRF and SJLR compute edge curvatures with relatively high computational costs, one may prefer a cheaper rewiring method. Furthermore, both SDRF and SJLR are based on the adjusted curvatures. The potential of the original curvatures (i.e., those described in Eq.\eqref{ollivier_original} and Eq.~\eqref{forman_original}) with simpler formulations to indicate the OSM and OSQ problems remains unknown. Based on these motivations, \cite{nguyen2023revisiting} linked the OSM and OSQ problems with the Ollivier-Ricci curvature defined in Eq.\eqref{ollivier_original}, and show that just like the JLC used in \cite{giraldo2022understanding}, the original Ollivier-Ricci curvature also has the capability to indicate the OSM and OSQ problems. Accordingly, a graph rewiring approach named Batch Ollivier-Ricci Flow (BORF) is developed with a provably cheaper computational cost than SDRF and SJLR. Another significant contrast between BORF and SDRF, as well as SJLR, is the methodology employed for computing graph curvature. BORF infrequently computes graph curvature by sharing the optimal transport plan across batches, in contrast to SJLR, which necessitates repeated recalculations for every potential new edge. Through the batch-wise rewiring of edges, BORF introduces a consistent alteration throughout the entire graph. This strategy contributes to the preservation of the graph's topology and mitigates the risk of continuously rewiring a small outlier subgraph, leaving other parts of the graph devoid of any geometric enhancements.

\paragraph{$\mathbf{AFR}$-3 \cite{new_fesser2023mitigating}}
To further reduce the computational cost from the aforementioned methods so that the curvature rewiring paradigm can be applied to large-scale benchmark datasets, \cite{new_fesser2023mitigating} considered the Forman curvature (Eq.~\eqref{forman_original}) on the higher-order graph structures. Specifically, \cite{new_fesser2023mitigating} considered the Augmented Forman Ricci curvature (AFRC) that evaluates higher-order information encoded in cycles of order $\leq k$
(denoted as $\mathcal {AF}_k$). When $k =3$ and $k=4$, the corresponding AFRC takes the form as\footnote{We note that to align the notation with the original paper, in the sequel, we use $AFRC$ and $AFR$ interchangeably.} 
\begin{align}
    \mathbf{AFR}_3(i,j) &\coloneqq 4- d_i - d_j + 3|\sharp_\Delta(i,j)|, \notag \\ 
    \mathbf{AFR}_4(i,j) &\coloneqq 4- d_i - d_j + 3|\sharp_\Delta(i,j)|+2 |\sharp_\Box(i,j)|,
\end{align}
where $|\sharp_\Delta(i,j)|$ denote as the number of triangles that contain edge $(i,j)$ and $|\sharp_\Box(i,j)|$ denotes the number of quadrangles that contain edge $(i,j)$, \revision{and the complexity of $\mathbf {AFR}_3(i,j)$ and $\mathbf {AFR}_4(i,j)$ are the same as $F(i,j)$ and $\mathrm{BFC}(i,j)$. See Table~\ref{summary_table} for a summary of all computational complexities of the listed curvatures.}
It is important to note that the definition of AFRC is different from the BFC defined in \cite{topping2021understanding}, and the derivation of $\mathbf{AFR}$(s) is available in \cite{forman2003discrete,new_tian2025curvature}. However, similar to BFC, once the graph becomes dense, i.e., more triangles and quadrangles contain edge $(i,j)$, both BFC and AFRC approach to their upper bound\footnote{For example, $\mathbf{AFR}_3$ is bounded by $4-d_i-d_j \leq \mathbf {AFR}_3 \leq d_j +1$, for node $i,j$ with $d_i \geq d_j$ \cite{new_fesser2023mitigating}.}, and OSM problem is alleviated.

Under the same strategy with BORF and SJLR, i.e., adding edges around edges with lower curvatures (for OSQ) and removing edges for highly curved edges (for OSM), the improvement from $\mathbf{AFR}$-3 can be summarized as follows.
\begin{itemize}
    \item Unlike BORF and SJLR which determine the number of added/removed edges by costly hyperparameter tuning, $\mathbf{AFR}$-3 automatically computes these numbers by using node-wised  Gaussian mixture model proposed in \cite{new_fesser2024augmentations}.

    \item The computational complexities of both $\mathbf{AFR}$-3 and $\mathbf{AFR}$-4 are less than those in other curvature rewiring methods.
\end{itemize}

In addition to the above curvature-based rewiring methods, we highlight that many attempts have been made recently by leveraging curvature information without adding (only dropping) edges to the graph, such as curvature-based edge drop \cite{liu2023curvdrop,shi2023curvature} and graph classification \cite{new_sanders2023curvature}, \revision{or treating curvature as an indicator to show effectiveness of the rewiring method \cite{attali2024delaunay}.}
\revision{Some related works \cite{sun2023deepricci,yang2023kappahgcn} learn curvature-aware representations in Riemannian spaces and use the resulting curvature signals to widen information bottlenecks, thus strengthening long-range message passing. Instead of rewiring, \cite{fesser2024effective} proposes to encode local discrete Ricci curvature at each node as features, which allows GNNs to adapt propagation to local geometry and, in experiments, often outperforms curvature-based rewiring methods that target over-squashing. \cite{DBLP:conf/bigdataconf/SunHCWL23} injects Ricci curvature into the contrastive learning objective so that the negatively curved (bottleneck) edges are penalized in the representation space.}


\revision{
\textit{Reflection on Curvature-based Methods \cite{new_tori2025the}}.
Although curvature based methods has identified the relationship between curvature and OSQ, and show exceptional learning performances in their empirical verifications, it is still unclear,  at least in empirical studies, to see whether these empirical gains are actually due to the effect from curvature based rewiring. One recent study \cite{new_tori2025the} shows that those edges
selected during the rewiring process are not aligned with the theoretical criteria for identifying bottlenecks.  For example, in practice, most of the edges that are actually marked for OSQ in SDRF \cite{topping2021understanding} satisfy a much looser selection condition (see Eq.~(9) and (10) in \cite{new_tori2025the}) compared to its theoretical counterpart, especially in the citation graphs, suggesting a limited practical effectiveness of the existing curvature-rewiring methods. In addition, \cite{new_tori2025the} also shows that the accuracy gain from SDRF in terms of long-range learning tasks can also be achieved by a thorough hyperparameter tuning, suggesting a limitation on how to isolate the effect of OSQ-mitigating methods empirically. We show more discussion on this matter in the opening questions in Section \ref{open_questions}.

}
\revision{
\paragraph{Layer-dependent Methods \cite{gutteridge2023drew,new_barbero2023locality}}

The fundamental concept behind the Dynamic Rewiring with Delay (DRew) proposed in \cite{gutteridge2023drew} is evident from its designation as \textit{dynamic}. This is a rewiring method that is layer-dependent, i.e., on each iteration of MPNN, DRew provides different rewiring results (thus resulting in different graphs) for feature propagation. In DRew (without delay), two nodes with distance $r$ are connected directly without considering letting their information travel through their intermediate neighbors. We emphasize that the MPNN with this design delves into the intermediate space between the original graph connectivity and the fully connected graph commonly utilized by graph transformers (See discussion in Section~\ref{sec:implicit_rewiring}).
Furthermore, the delay mechanism is also introduced to DRew so that, on each iteration of GNNs, the generated node feature depends on the set of the current and historical features, yielding a distance-aware skip-connection type rewiring scheme.  

Similar to the motivation for designing a dynamic graph rewiring method,  \cite{new_barbero2023locality} pointed out that one ideal rewiring approach shall be able to: (i) reduce OSQ, (ii) respect the locality of the graph, and (iii) preserve the global sparsity of the graph. \cite{new_barbero2023locality} further highlighted the trade-off between spatial- and spectral-based rewiring methods. We note that spatial methods often satisfy (i) and (ii) but not (iii) whereas the spectral methods can handle (i) and (iii) at the expense of (ii). Accordingly, LASER is proposed to satisfy all of (i)-(iii) through a locality-aware sequence of rewiring operations. More specifically, to balance the local connectivity and meanwhile mitigate the OSQ problem, two measures, namely locality ($\pi_L : \mathcal V\times \mathcal V \rightarrow \mathbb R$) and connectivity ($\pi_C :\mathcal V\times \mathcal V \rightarrow \mathbb R$) are defined with $\pi_L$ measuring how well nodes are connected in the graph and $\pi_C$ showing the quantity that penalizes interactions among nodes that are distant according to some metric on the input graph. More specifically, in LASER \cite{new_barbero2023locality}, $\pi_C$ is chosen as $\pi^{(\ell)}_C (i,j) : = (\widehat{\mathbf{A}}^{\ell})_{i,j}$,
which equals the number of walks from $i$ to $j$ of length at most $\ell$. We highlight that, in case of any confusion, here $\widehat{\mathbf{A}}^{\ell}$ stands for $\widehat{\mathbf{A}}$ raised to the power $\ell$. It is straightforward to see that, if $\pi^{(\ell)}_C (i,j)$ is large, then two nodes $i,j$ have multiple alternative routes to exchange information and would usually have small effective resistance\footnote{One can also measure this through other indicators such as Cheeger constant, curvature, spectral gap and commute time.}. On the other hand, $\pi_L$ is chosen as the standard shortest distance $d_\mathcal G$, and, in particular, LASER further restricts that $d_\mathcal G = \ell +1$. This suggests that, at layer $\ell$, the LASER only adds edges among nodes at a distance exactly $\ell +1$. Lastly, the LASER rewiring method is conducted by rewiring those edges with the lowest connectivity score ($\pi^{(\ell)}_C$) to a set of nodes with fixed shortest distance. Moreover, it can be demonstrated that LASER tends to rewire the graph in a manner that better preserves locality, whereas global spectral approaches often result in the loss of information contained in the locality of the original graph \cite{new_barbero2023locality}.

}

\revision{
\paragraph{Virtual Nodes Methods \cite{geisler2024spatio} \cite{qian2024probabilistic, shi2025design,gabrielsson2023rewiring}}
Virtual node methods address the OSQ problem by either 
explicitly adding supporting nodes to the graph so that richer connections can be generated between the original graph nodes and virtual nodes \cite{shi2025design,qian2024probabilistic,gabrielsson2023rewiring}  or by showing that the feature propagation of the model is equivalent to having virtual nodes in the graph \cite{geisler2024spatio}. In \cite{geisler2024spatio}, the spatial and spectral feature propagations are combined such that 
\begin{align}\label{s2gnn}
    \mathbf H(\ell+1) = \mathrm{Spectral}\left(\mathbf H(\ell); \mathbf U, \boldsymbol{\Lambda}\right) + \mathrm{Spatial} \left((\mathbf H(\ell); \mathbf A\right),
\end{align}
where $\mathrm{Spectral}(\cdot)$ and $\mathrm{Spatial}(\cdot)$ stand for any spectral filtering and spatial propagations in GNNs. It can be shown that $S^2$GNN propagation defined in Eq.~\eqref{s2gnn} is equivalent to adding virtual nodes to the graph. This is because the first eigenvector calculates the average feature embedding and then distributes this information throughout the whole dynamics, while other eigenvectors effectively enable messages to be passed within or between graph clusters. Therefore, from the spatial dynamic ($\mathrm{Spatial}(\mathbf H; \mathbf A)$) point of view, the spectral dynamic provides additional information between distant nodes that are outside the current receptive field. More importantly, a \textit{lower bound} of the OSQ score in Proposition \ref{OSQ_topping} is as follows.
\begin{lem}[\cite{geisler2024spatio}]\label{lem:s2-lower}
Consider an $\ell$-layer S$^2$GNN as defined in Eq.~\eqref{s2gnn}. Let $(\tilde{\vartheta},\vartheta)$ be, respectively, the parameters of the spatial
and spectral parts. Assume there exist $\tilde{\vartheta}$ such that
$\mathrm{Spatial} \,\big(\mathbf H{(\ell)}; \widehat{\mathbf A},\tilde{\vartheta}\big)=\mathbf 0$
for all $1\le \ell\le r$, and $f_\theta=\mathrm{Id}$. Then one can choose
spectral filters $\vartheta$ so that the network expresses a signal
$\mathbf h_i^{(r+1)}$ whose
Jacobian sensitivity admits a uniform positive lower bound:
\begin{equation}\label{eq:lower-L1}
\left\Vert
\frac{\partial \mathbf h_{i}^{(r+1)}}{\partial \mathbf x_s}
\right\Vert
\;\ge\;
\begin{cases}
\dfrac{w\,K_\vartheta^{r+1}}{2\,|\mathcal E_C|}, & \text{if } i\sim s,\\[6pt]
0, & \text{otherwise,}
\end{cases}
\end{equation}
where $C$ is a connected component in graph, $w$ is the width (number of channels of $f_\theta$), $|\mathcal E_C|$ is the number of edges in $C$, and $K_\vartheta>0$ is a constant determined by the filter choice.
\end{lem}
We note that we did not include this Lemma in \ref{OSQ_bounds_1} because all conclusions in Section \ref{OSQ_bounds_1} present the bounds of OSQ based on different facets (e.g., effective resistance and commute time) under a fixed MPNN propagation in Eq.~\eqref{mpnn}, whereas the conclusion in Lemma \ref{lem:s2-lower} is derived based on the propagation in Eq.~\eqref{s2gnn}. It can be observed from Lemma \ref{lem:s2-lower} that $S^2$GNN vanquishes the OSQ problem, i.e., $\frac{w\,K_\vartheta^{r+1}}{2\,|\mathcal E_C|}>0$ for all nodes that belong to the same cluster. A further result is derived in a recent successor, namely $S^3$GNN under relaxed conditions \cite{shi2026s}. 

On the other hand, methods like IPR-MPNNs \cite{qian2024probabilistic} and UYGNNs \cite{shi2025design} explicitly add virtual nodes to the graph, with the former depending on an assignment matrix learned from its upstream model connected by $k$ edges and the latter depending on the variations of the labels in the training set. These virtual nodes with their edges enrich the connectivity of the graph so that OSQ problem is mitigated. Recently, a thorough analysis on the effects of virtual nodes was conducted in \cite{southern2025understanding}. Different from UYGNN \cite{shi2025design} which builds connections based on node labels,  \cite{southern2025understanding} connects all virtual nodes to all other nodes, and we summarize their main findings as follows. 

\begin{lem}[\cite{southern2025understanding}]\label{lem:commute}
  Let $\mathrm{Com}(i,j)$ be the commute time between node $i$ and $j$, then if one virtual node is inserted and connected to all other nodes, the changes of the commute time (denoted as $\Delta\mathrm{Com}(i,j)$) is 
  \begin{equation}\label{eq:commute-change}
    \resizebox{0.49\textwidth}{!}{$
    \begin{aligned}
        \Delta\mathrm{Com}(i,j)
        = 2|\mathcal E| \sum_{i'=2}^{N}
        \frac{1}{\lambda_{i'}(\lambda_{i'}+1)}
        \left(\frac{N}{|\mathcal E|}\lambda_{i'} - 1\right)
        (\mathbf u_{i'}(i) - \mathbf u_{i'}(j))^2
    \end{aligned}
    $}
    \end{equation}
In addition, the average change of the commute time is 
\begin{equation}
    \resizebox{0.4\textwidth}{!}{$
    \begin{aligned}
        \frac{1}{N^2}\sum_{i,j}^N \Delta\mathrm{Com}(i,j) = 
    \frac{4|\mathcal E|}{N}\sum_{i'=2}^{N}\frac{1}{\lambda_{i'}(\lambda_{i'}+1)}\left(\frac{N}{|\mathcal E|}\lambda_{i'} -1\right).
    \end{aligned}
    $}
    \end{equation}
\end{lem}


We note that our Laplacian index ranges from $[1, N]$, whereas in \cite{southern2025understanding} it starts from 0. The proof of Lemma~\ref{lem:commute} is based on the effective resistance in Eq.~\eqref{er_formular}, and its conclusions extend to the commute time since they are proportional. \cite{southern2025understanding} empirically shows that adding a virtual node decreases the commute time between inter-community nodes—thus mitigating the OSQ problem—though it may increase commute times within communities.

In addition, it is worth noting that recent works  \cite{hariri2025return,hariri2024graph,shi2024graph,jiang2024limiting} also verify that the spectral filtering paradigm \footnote{The term spectral filtering (feature propagation) differs from the spectral rewiring methods (spectral indicator for rewiring) introduced in Section \ref{sec:spectral_rewiring}.}, such as ChebNet, naturally has an advantage in mitigating the OSQ problem due to its global view of the whole graph via the spectral domain. Building on this advantage, heat kernel-based methods \cite{shao2023unifying,new_bamberger2025bundle} have been proposed, with the former mitigating the OSQ problem via multi-scale graph convolution and the latter deriving bundle neural networks layer by assigning each node a vector space and orthogonal map, resulting in a kernel-based heat diffusion process. Furthermore, an universal polynomial filter is developed in \cite{huang2024universal} and shown that in the propagation induced from such a filter, the OSQ score is independent of the layer number of $\ell$.

\begin{tcolorbox}[breakable]
\textbf{Takeaway Message of this Section:} \textit{\revision{Spatial rewiring methods mitigate the OSQ problem by exploiting local topological indicators (e.g., graph curvature), augmenting the graph with additional structural information (e.g., virtual nodes), supplying additional connectivities within feature propagation (e.g., dynamic rewiring), or incorporating global dependencies between distanced nodes (e.g., through spectral filtering).}}
\end{tcolorbox}

}

\subsection{Spectral Rewiring Techniques}\label{sec:spectral_rewiring}
\revision{
Spectral rewiring usually builds edges between nodes so that some graph global expansion properties (i.e., graph global indicators), such as spectral gap \cite{new_karhadkar2023fosr}, total effective resistance \cite{black2023understanding} and commute time \cite{new_di2024does}, are optimized.  
}


\paragraph{Expander Graph Propagation (EGP) \cite{deac2022expander}}
\cite{deac2022expander} proposed the so-called expander graph in addition to the classic MPNN propagation for the OSQ problem. Specifically, by leveraging the special linear group $\mathrm{SL}(2,\mathbb Z_n)$, one can construct a family of expander graphs (i.e., Cayley graphs). These graphs provide additional connectivity information through the propagation of message-passing neural networks (MPNNs). Formally, EGP-based GNN model can be defined as
\begin{align}\label{EGP}
    \mathbf{H} = \mathrm{GNN} (\mathrm{GNN}(\mathbf X, \mathbf{A}), \mathbf{A}^{\mathrm{Cay(n)}}),
\end{align}
where $\mathbf{A}^{\mathrm{Cay(n)}}$ is the adjacency matrix generated from the Cayley graph family. In addition, the link between the local structure of the Cayley graphs and BFC \cite{topping2021understanding} as well as the Ollivier-Ricci curvature (Eq.~\eqref{ollivier_original}) has also been discussed. Unlike other rewiring methods that directly modify the input graph adjacency, EGP enhances the node features generated by the base GNN model by providing supplementary adjacency information, acting as a supportive graph. It is noteworthy that EGP exhibits a subquadratic complexity of $\mathcal O(\mathrm{log}(N))$, making it relatively more cost-effective compared to other rewiring methods. \revision{It should be noted that, motivated similarly to EGP, a recent study \cite{new_wilson2024cayley} propagates feature information over a complete Cayley graph structure to address the potential information loss in EGP, where the computation graph is truncated to align with the given input graph.}

\paragraph{Total Effective Resistance Rewiring \cite{black2023understanding}}
In addition to utilizing curvatures as indicators of the OSQ problem, \cite{black2023understanding} further measures the OSQ problem through the concept of effective resistance. We recall that for any node pair $i$ and $j$ of the graph, its effective resistance is formulated as 
$R_{i,j} \coloneqq \left(\frac{1}{\sqrt{d_i}}\boldsymbol{1}_i -\frac{1}{\sqrt{d_j}}\boldsymbol{1}_j\right)^\top \widehat{\mathbf L}^+  \left(\frac{1}{\sqrt{d_i}}\boldsymbol{1}_i -\frac{1}{\sqrt{d_j}}\boldsymbol{1}_j\right)$, where $\widehat{\mathbf L}^+$ is the pseudoinverse of $\widehat{\mathbf L}$ and $\boldsymbol{1}_i$, $\boldsymbol{1}_j$ are the indicator vectors of node $i$ and $j$, respectively. It is evident that mitigating the OSQ problem necessitates a preference for a rewired graph with the minimum sum of effective resistances, denoted as $R$. Based on this observation, a graph rewiring method is developed by adding edges to the graph so that $\sum_{i,j \in \mathcal V} R_{i,j}$ is minimized. Additionally, one variant of the aforementioned rewiring method that maximizes $B^2_{i,j}/(1+R_{i,j})$ is  proposed namely greedy total resistance rewiring (GTR rewiring), where $B_{i,j} : = \sqrt{\left(\boldsymbol{1}_i -\boldsymbol{1}_j\right)^\top (\widehat{\mathbf L}^+)^2  \left(\boldsymbol{1}_i -\boldsymbol{1}_j\right)}$,
is the biharmonic distance between node $i$ and $j$ \cite{black2023understanding}. In addition, we point out that the rewiring scheme based on effect resistance is also developed via Lov\'asz bound in \cite{new_arnaiz2022diffwire}. 

\paragraph{Random Local Edge Flip (RLEF) \cite{banerjee2022oversquashing}}
Most of the curvature rewiring methods mentioned above carry the risk of causing graph disconnection. These methods typically involve deleting edges, particularly those associated with highly positive curvature. Although this situation can be prevented by careful hyperparameter tunning \cite{topping2021understanding}, a more powerful rewiring algorithm with a theoretical guarantee to maintain graph connectivity is desirable. In  \cite{banerjee2022oversquashing}, this goal is achieved by maintaining the Cheeger constant which measures how well two sets of nodes $\mathcal T_1$ and $\mathcal T_2$ are connected, whereas the curvature rewiring methods can be treated as the methods that maintain local Cheeger constant. One can see that if the initial input graph is connected, with the preservation of the Cheeger constant during rewiring, the graph will never become disconnected. Building upon this, the RLEF rewires the graph through the Flip Markov chain, which performs an edge flip that amounts to exchanging a random neighbor. More specifically, the  Flip Markov chain selects a random three-path $i-u-v-j$ before executing the edge flip. This process involves initially choosing a hub edge $(u, v)$ uniformly at random. Subsequently, $i$ and $j$ are selected as random neighbors of $u$ and $v$, respectively, with the condition that all four nodes must be distinct. Accordingly, RELF transformation preserves the degree of all nodes as well as their connectivity. Notably, the RLEF method for addressing the OSQ problem is inspired by concepts of information contraction and graph expansion. This introduces a novel perspective on the OSQ issue, distinct from traditional approaches involving curvature and effective resistance.

\paragraph{First Order Spectral Rewiring (FOSR) \cite{new_karhadkar2023fosr}}
As RLEF is proposed with the purpose of handling the original topology of the graph to prevent it from being disconnected. It is natural to consider another adverse effect of graph rewiring that is potentially making the graph too dense. Recall that a once densely connected graph incurs the OSM problem. Therefore, an ideal rewiring method in this view shall be able to mitigate the OSQ problem while also preventing the OSM problem. To achieve this goal, FOSR attempts to maximize the spectral gap\footnote{A graph is regarded as having a good spectral expansion (measured by the Cheeger constant) if it has a large spectral gap \cite{new_karhadkar2023fosr,topping2021understanding}.}, which is the value of the second smallest eigenvalue (i.e. $\lambda_2$) of $\widehat{\mathbf L}$. Given we have $\widehat{\mathbf L} = \mathbf{I} - \widehat{\mathbf{A}}$, this is equivalent to maximizing the second-largest eigenvalue of $\widehat{\mathbf{A}}$. In light of this motivation, each iteration of FOSR involves adding an edge to each node (i.e., node $i$) in the graph. This addition is done by minimizing $\frac{2 \boldsymbol{u}_2[i] \boldsymbol{u}_2[j]}{\sqrt{(1+d_i)(1+d_j)}}$, representing the first-order change of the second-largest eigenvalue of $\widehat{\mathbf{A}}$. Remarkably, FOSR does not require to have the full eigendecomposition of either $\widehat{\mathbf{A}}$ or $\widehat{\mathbf L}$. Instead, it only requires an approximation of the second eigenvalue of $\widehat{\mathbf L}$ using power iterations, resulting in significant computational complexity savings. Lastly, FOSR adapts a relational GNN paradigm by assigning different labels to the added edges and propagating node features separately via original and newly rewired edges to mitigate the OSM problem.

\revision{
\paragraph{Spectrum-Preserving Sparsification (GOKU) \cite{liang2025mitigating}} GOKU \cite{liang2025mitigating} proposes a two-step Densification-Sparsification Rewiring (DSR) pipeline that explicitly preserves the (Laplacian) spectrum. Specifically, GOKU produces a rewired graph with Laplacian $\tilde{\mathbf L} \in \mathbb R^{n\times n}$ satisfying $(1-\epsilon) \mathbf x^\top \mathbf L \mathbf x \leq  \mathbf x^\top \tilde{\mathbf L} \mathbf x \leq (1+ \epsilon) \mathbf x^\top \mathbf L \mathbf x$
for any $x \in \mathbb R^n$. This allows the smallest Laplacian eigenvalues to be increased without significantly affecting the remaining spectrum. This balances the reduced over-squashing without increasing the risk of over-smoothing relative to the input graph.  
}

\begin{tcolorbox}[breakable]
\textbf{Takeaway Message of this Section:} \textit{Spectral rewiring focuses on optimizing the global graph expansion properties such as the spectral gap, total resistance, and commute time, allowing the algorithm to add edges between non-local nodes. {Compared with spatial rewiring, it better targets non-local bottlenecks but may weaken locality preservation.}
}
\end{tcolorbox}



\subsection{Implicit Rewiring}\label{sec:implicit_rewiring}
In this section, \revision{we summarize three groups of implicit rewiring methods, namely diffusion-based rewiring methods, graph transformer-based methods, and multi-track feature propagation methods. We note that, although these approaches may not primarily target the OSQ problem or may not involve explicit rewiring, the feature propagation rules in these models can be equivalently regarded as updating node features over a "better-connected" graph (e.g., denser, learnable or fully-connected) or through carefully designed channels, thereby constituting an implicit rewiring paradigm \cite{dwivedi2022long}. Although not empirically verified, these methods are expected to yield improved performance on long-range tasks compared to conventional GNNs (e.g., GCNs). This is because the OSQ problem originates from insufficient information exchange between nodes imposed by the graph topology; thus, in principle, a fully-connected graph is inherently less susceptible to OSQ in standard MPNN architectures.
}
We start with the diffusion-based methods.  


\subsubsection{Diffusion Inspired Message Passing}

\paragraph{DIGL \cite{gasteiger2019diffusion}}
Even before \cite{topping2021understanding} established the link between the OSQ problem and graph curvature, a graph diffusion model, known as the graph diffusion convolution (GDC) \cite{gasteiger2019diffusion}, was developed. This model leveraged the graph heat kernel and personalized PageRank to alleviate issues arising from arbitrarily defined edges \revision{ (e.g., some of them may be responsible for the OSQ problem) } in the graph. Specifically, DIGL sparsifies the graph adjacency matrix by either using $k$-entries with the highest mass per column or setting entries to 0 if they are below a specific threshold. We highlight that, although the sparsification in DIGL is relatively simple, the computational complexity of DIGL is as low as $\mathcal O(N)$. In fact, DIGL now serves as a standard baseline rewiring method in many recent studies \cite{new_karhadkar2023fosr,black2023understanding}, and as the foundational model for the general graph diffusion models defined as follows. 

\paragraph{Graph Neural diffusion (GRAND) \cite{chamberlain2021grand} and Beltrami Diffusion \cite{chamberlain2021beltrami}}
The aforementioned strategies utilize graph structure, i.e., the various forms of curvature, to rewire the graph. Recent works have also exploited the learned features to achieve the same purpose \footnote{We note that although one may categorize diffusion methods \cite{chamberlain2021grand,chamberlain2021beltrami} as into implicit rewiring methods, given their closed link to DIGL, for illustration purposes, we list them here for self-completeness, and more detailed discussion on OSQ problem can be found in Section \ref{sec:implicit_rewiring}. }. In particular, graph neural diffusion (GRAND) \cite{chamberlain2021grand} and Beltrami diffusion (BLEND) \cite{chamberlain2021beltrami} are GNNs that perform graph rewiring inspired by the continuous dynamics \cite{new_han2024from,gasteiger2019diffusion}.
The propagation of GRAND encodes the anisotropic diffusion on graphs via attention, i.e.,
\begin{align}
    \frac{\partial \mathbf{H}}{\partial t} = \div \big( \mathbf{A}(\mathbf{H}) \odot \nabla \mathbf{H} \big), \label{eq:grand}
\end{align}
where $\mathbf{A}(\mathbf{H}) \in \mathbb R^{N\times N}$ is the diffusion coefficients computed by attention. In GRAND,  $\mathbf{A}(\mathbf{H})_{i,j} = 0$ if $(i,j) \notin \mathcal E$. We denote $\odot$ as the element-wise product between two matrices and $\nabla$ is the graph gradient that collects the node feature difference along edges. When $\mathbf{A}(\mathbf{h}_i, \mathbf{h}_j) = \mathbf{A}$ is fixed for the entire diffusion process, the above dynamics reduces to graph heat diffusion as $\frac{\partial \mathbf{H}}{\partial t} = \div \big( \mathbf{A} \odot \nabla \mathbf{H} \big) = -\mathbf L\mathbf{H}$, in which $\mathbf L = \mathbf D - \mathbf{A}$. 
In GRAND, graph rewiring can be performed based on the learned diffusion coefficients dynamically across the diffusion process. More specifically, the graph is gradually densified as $\mathcal E \leftarrow \mathcal E\cup \{ (i,j):  \mathbf{A}(\mathbf{H})_{i,j} > \xi \}$ for some threshold $\xi > 0$. This follows the work in \cite{gasteiger2019diffusion} 
where each runtime the subset of edges to use is learned based
on attention weights, serving as a combination of pre-processing and dynamic rewiring processes. Although without theoretical justification, i.e., on the OSQ problem, it is observed that GRAND with rewiring generally improves the performance of vanilla GRAND, indicating promise for conducting a discrete diffusion process on the rewired graph. 

Graph Beltrami Diffusion (BLEND) extends the framework of GRAND by augmenting input features $\mathbf{h}_i$ with their corresponding positional coordinates $\mathbf u_i$ for each node. If the positional information is not available, embedding methods can be used, such as personalized PageRank \cite{gasteiger2019diffusion}, deepwalk \cite{perozzi2014deepwalk}, and hyperbolic embeddings \cite{chami2019hyperbolic}. The diffusion process is equivalent to GRAND \eqref{eq:grand} after setting $\mathbf{H} \leftarrow [\mathbf{H}, \mathbf U]$.
The dynamics of BLEND is motivated by Beltrami flow on Riemannian manifolds, in which the Laplace-Beltrami operator is position-dependent due to the changing Riemannian metric. Similar to GRAND, graph rewiring can be leveraged to enhance the dynamics of BLEND. Compared to GRAND, BLEND can explicitly utilize the positional information to rewire the graph based on the spatial proximity as $\mathcal E \leftarrow { (i,j) : d_\mathcal C(\mathbf u_i, \mathbf u_j) < \xi }$ or by the k-nearest neighbor graph. Here $d_{\mathcal{C}}(\cdot, \cdot)$ 
 is a distance function for the metric space of the nodes. \revision{A more recent work \cite{huang2025torque} presents a rewiring technique based on a physics-inspired definition of torque per edge. It measures how neighbor’s message interferes with a node’s representation and proposes to prune edges with high torque and add edges with low torque in order to enhance message passing efficiency and reduce over-squashing. \cite{ribeiro2025cooperative} adopts a similar strategy by allowing  node actions to be learned, i.e., sending and receiving information based on Sheaf theory. This is equivalent to adding or removing edges and thus implicitly addresses over-squashing.}

\paragraph{Fractional Diffusion \cite{maskey2023fractional}}
So far the dynamics in our reviewed models mainly focus on the local information of nodes, which intuitively can be reckoned as propagating node features via node local neighboring structures.
This observation also supports that the rewiring paradigm within GRAND and BLEND is helpful for the model to capture the long-term dependencies between nodes, therefore potentially mitigating the OSQ problem. Based on this observation, some non-local dynamic-based MPNNs have been recently developed.  For instance, the recent work \cite{maskey2023fractional} introduces the fractional graph Laplacian/adjacency matrix, denoted as $\widehat{\mathbf{A}}^\tau$ with $\tau \in \mathbb R$, giving rise to a non-local diffusion process termed Fractional Laplacian Neural ODE (FLODE). Formally, it is defined by the propagation rule $\frac{\partial \mathbf{H}}{\partial t} = -\widehat{\mathbf{A}}^\tau \mathbf{H}$, where the preference for a fractional value of $\tau$ results in denser counterparts for $\widehat{\mathbf{A}}^\tau$ and $\widehat{\mathbf L}$ compared to their original forms. By adjusting the value of $\tau$, it has become possible to propagate feature information through the graph at varying levels of connectivity. Consequently, it is evident that the rewired graph indirectly induced by FLODE exhibits increased robustness against the OSQ problem, \revision{ i.e., through a denser graph adjacency.} 
\revision{Another work \cite{new_eliasof2025tango} introduces a tangential direction to the gradient flow of the Dirichlet energy, which approximates the Newton update. This improves the conditioning of the Hessian of energy, and thus serves as a surrogate for mitigating over-squashing.}

\paragraph{Hypo-elliptic Diffusion \cite{toth2022capturing}}
Another path of enforcing the graph diffusion model to mitigate the OSQ problem is to re-propagate the features from the previous propagation results, yielding a re-utilization of the neighboring information. Hypo-elliptic Diffusion achieves this goal by treating the feature propagation results as $\mathbf{h} = (\mathbf{h}^{(0)}, \mathbf{h}^{(1)}, \cdots \mathbf{h}^{(\ell)}) \in (\mathbb R ^{c_0})^{\ell +1}$ as an element of the sequence space $\mathrm{Seq}(\mathbb R^c_0)$. Let $\varphi: \mathbb R^{c_0} \rightarrow \mathcal{H}$ be an algebra lifting, which allows to define a sequence feature map $\tilde{\varphi} (\mathbf{h}) 
= \varphi(\mathbf{h}^{(0)} \cdot \varphi(\mathbf{h}^{(1)} - \mathbf{h}^{(0)} \cdots \varphi(\mathbf{h}^{(\ell)} - \mathbf{h}^{(\ell-1)}) \in \mathcal{H}$, where $\mathcal{H}$ is the product Hilbert space. The corresponding diffusion process requires a tensor adjacency matrix, $\widetilde{\mathbf{A}} \in \mathcal{H}^{N \times N}$ with the entries $\widetilde{\mathbf{A}}_{i,j} = \varphi(\mathbf{h}_j^{(0)} - \mathbf{h}_i^{(0)}) \in \mathcal{H}$ if $(i,j) \in \mathcal E$ and $0$ otherwise. The associated Laplacian $\widetilde{\mathbf L}$ can be defined accordingly. For example, the random walk Laplacian has entries $\widetilde{\mathbf L}_{i,i} = 1$ and $\widetilde{\mathbf L}_{i,j} = - \frac{1}{d_i} \varphi(\mathbf{h}_j^{(0)} - \mathbf{h}_i^{(0)})$ if $(i,j) \in \mathcal E$ and $0$ otherwise. It is not hard to check that $\widetilde{\mathbf L}$ in this format is encoded with the feature historical information. The corresponding hypo-elliptic graph diffusion can be defined as $\frac{\partial \tilde{\varphi}(\mathbf  H)}{\partial t} = - \widetilde{\mathbf L}  \tilde{\varphi}(\mathbf{H})$. One can see that propagating node features together with their histories is equivalent to connecting all nodes together with different edge weights according to the functional $\tilde{\varphi}$. \revision{ Thus such a method has the potential of mitigating the OSQ problem as stated in \cite{toth2022capturing}, in which the authors claimed that the hypo-elliptic graph Laplacian provides a structured way to capture long-range dependencies on graphs and is robust to pooling, making the method less susceptible to the over-squashing phenomenon.}

\paragraph{Quantum Diffusion Convolution \cite{new_markovich2024qdc}}
With a similar spirit of incorporating the relationship between node features in diffusion process, \cite{new_markovich2024qdc} defines a spectral filtering approach that acts on the average overlaps of node features, serving as the quantum diffusion kernel. Specifically, the average overlap between any two nodes $i,j$ is defined as the inner product between $| \psi(t) \rangle_i$ and $| \psi(t) \rangle_j$, denoted as $\sum_{k,l = 1}^n c_k^* c_l | \boldsymbol{\phi}_k \rangle_i^* | \boldsymbol{\phi}_l \rangle_j$, which is analogized as the overlap of two quantum states (probability densities of two quantum particles) between node $i$ and $j$ at time $t$, respectively. The paper further leverages a Gaussian filter $\mathcal P = \sum_{k=1}^n \mathrm{e}^{- (\lambda_k - \mu)^2/{2 \sigma^2}}$, where $\lambda_k$ is the $k$-th eigenvalue of the graph Laplacian, and the corresponding quantum diffusion kernel (QDC) is defined as  $\mathbf Q_{i,j} = \sum_{k=1}^n e^{- (\lambda_k - \mu)^2/{2 \sigma^2}} |\boldsymbol{\phi}_k\rangle_i^*  |\boldsymbol{\phi}_k\rangle_j$. Similar to the Hypo-elliptic Diffusion, when QDC is used as a transition matrix via diffusion, the relative relationship between nodes is considered. Therefore, for each iteration, all pairs of feature similarities are involved via message passing, suggesting a potential alleviation of the OSQ problem.

\subsubsection{Graph Transformers }
Transformer \cite{vaswani2017attention} and its variants compute attention scores over fully connected or grid graphs (that are much denser), which achieve state-of-the-art performance. Motivated by its success, many recent developments consider applying the attention paradigm of transformer to MPNN models, leading to a class of graph transformer models \cite{ying2021transformers,new_dwivedi2021generalization,rampavsek2022recipe,chen2022structure,he2023generalization,new_muller2024attending}. The key process of the transformer-based graph models is to assign attentions for all node pairs regardless of whether they are originally linked or not (known as global receptive field) so that message passing can be conducted through a properly \textit{rewired} graph. We note that as the ability of transformer-based models to mitigate the OSQ problem has been widely observed empirically \cite{new_barbero2023locality,dwivedi2022long}, we only included several of them for illustration purposes. 



\paragraph{Graphormer \cite{ying2021transformers}}
Graphormer assigns attention to the graph adjacency matrix via the so-called spatial and edge encoding process. Specifically, similar to the classic transformer, node features are firstly encoded via the Query-Key encoding scheme, that is $\mathbf Q = \mathbf{H}\mathbf W_Q, \mathbf K = \mathbf{H} \mathbf W_K,$
in which $\mathbf W_Q$ and $\mathbf W_K\in \mathbb R^{c_0\times d_K}$ are the initial projections of the node features. Node features of the graph may also be augmented by the relative importance in degree (i.e., centrality). Specifically, for each input feature vector $\mathbf x^0_i$, one can construct $\mathbf{h}^{(0)}_i = \mathbf x^{(0)}_i + \mathrm{deg}^-(i) + \mathrm{deg}^+(i)$, where $\mathrm{deg}^-(i)$ and $\mathrm{deg}^+(i) \in \mathbb R^{c_0}$ are learnable embedding vectors determined by the in and out-degree of node $i$, respectively. 
After the initial encoding phase, the graph adjacency matrix can be enriched by the \textit{spatial encoding} and \textit{edge encoding}. Specifically, Graphormer considers a function $\zeta(i, j): \mathcal V \times \mathcal V \rightarrow \mathbb R$ for the node $i$, $j \in \mathcal V$\footnote{In case of any unclear, the notation $(i,j)$ here stands for the node pair $i$ and $j$, instead of edge $(i,j)$ as defined for curvatures.},
which measures the spatial relationship, which is chosen to be the short path distance if node $i$ connects to node $j$, and $-1$ otherwise. The result of $\zeta(i, j)$ serves as the bias term to the graph adjacency matrix, that is 
\begin{align}
    \mathbf{A}_{i,j} = \frac{(\mathbf{h}_i \mathbf W_Q)(\mathbf{h}_j\mathbf W_K)^\top}{\sqrt{d}} + b_{\zeta(i, j)} \,\,\, \forall i,j \in \mathcal V,
\end{align}
where $b_{\zeta(i, j)}$ is a learnable scalar indexed by $\zeta(i, j)$, and shared across all layers. The inclusion of $\zeta(v_i, v_j)$ for each node in a single Transformer layer can adaptively attend to all other nodes according to the graph structure information. More specifically, when $b_{\zeta(i, j)}$ is a decreasing function with respect to $\zeta(i, j)$, the mode will pay more attention to the lower degree neighboring information and less attention to the neighbors that far away from every initial node. In addition to the spatial encoding process, the edge encoding process in Graphormer further brings the edge information into the correlations between nodes. Specifically, for each ordered node pair $i, j$, Graphormer computes one of the shortest path $\mathbf {SP}(i,j) = (e_1, e_2,\cdots e_N) \subseteq d_\mathcal G(i,j)$ from $i$ to $j$, and compute an average of the dot-products of the edge feature and a learnable embedding along the path. Together with the spatial encoding, the final adjacency matrix can be formulated as 
\begin{align}
     &\mathbf{A}_{i,j} = \frac{(\mathbf{h}_i \mathbf W_Q)(\mathbf{h}_j\mathbf W_K)^\top}{\sqrt{d}} + b_{\zeta(i, j)} + c_{i,j} \,\,\, \notag \\
     &\mathrm{where} \,\,\, c_{i,j} = \frac{1}{N}\sum_{n=1}^N \mathbf x_{e_n}(\mathbf w_n^E)^\top,  \forall i,j \in \mathcal V,
\end{align}
where $ \mathbf x_{e_n}$ is the feature of the $n$-th edge $e_n$ in $\mathbf {SP}(i,j)$, $\mathbf w_n^E \in \mathbb R^{d_E}$ is the $n$-th weight embedding. Finally, with the newly designed graph adjacency, the Graphormer layer is built upon the self-attention (on the adjacency) and the feed-forward blocks (of layer normalization) over the node features \cite{ying2021transformers}.

Note that in the subsequent study of Graphormer \cite{shi2022benchmarking}, the robustness of the model on the OSQ problem is verified empirically via large-scale molecular datasets. In addition, the challenge of scalability of transformer-based MPNNs has also been explored in \cite{wu2022nodeformer}. Lastly, we highlight that recent research also attempted to assign positional information to enrich 
node features via propagation of MPNNs \cite{sun2022position} and transformers \cite{kreuzer2021rethinking}. In \cite{kreuzer2021rethinking}, two learned positional attentions, namely node and edge feature positional attentions, are generated via graph spectral information based on the initial assumptions on the graph spectra. A follow-up work \cite{mialon2021graphit} engages the node positional attentions based on positive definite kernels (i.e., graph heat kernel) on graphs,  enumerating and encoding local sub-structures such as paths of short length.

\paragraph{Difformer\cite{new_wu2023difformer}} 
In a recent work \cite{new_wu2023difformer}, the attention score function is defined to be more general compared with \cite{ying2021transformers} as $\widehat{\mathbf S}^{(\ell)}_{i,j} = 
\frac{f(\|\mathbf{h}^{(\ell)}_i - \mathbf{h}^{(\ell)}_j\|)}{\sum_{k = 1}^N f(\|\mathbf{h}^{(\ell)}_i - \mathbf{h}^{(\ell)}_k\|)}, \,\, 1\leq i,j \leq N$ for some non-negative
and decreasing function function $f$.
The Difformer model defines the message passing scheme as follows. 
\begin{equation}
\resizebox{0.49\textwidth}{!}{$
\begin{aligned}
    \mathbf{h}_i^{(\ell +1)} = \left(1-\frac{\tau}{2}\sum_{j=1}^N 
    \left(\widehat{\mathbf S}^{(\ell)}_{i,j} + \widehat{\mathbf{A}}_{i,j} \right)\right)
    \mathbf{h}_i^{(\ell)} + \frac{\tau}{2}\sum_{j=1}^N 
    \left(\widehat{\mathbf S}^{(\ell)}_{i,j} + \widehat{\mathbf{A}}_{i,j} \right)
    \mathbf{h}_j^{(\ell)}.
\end{aligned}
$}
\end{equation}
We further note that the inclusion of $\widehat{\mathbf{A}}_{i,j}$ is optional in Difformer when the model is required to incorporate more information from the original graph. Furthermore, follow-up research \cite{wu2023advective} has demonstrated the robustness of Difformer and its variants on the graphs with ground-truth labels are
independent of the observed graphs in data generation.


\revision{
\subsubsection{Multi-track Feature Propagation Methods \cite{pei2024multitrack,new_choi2024panda}}
Multi-track methods adjust (enhance) the quality of information flows in GNNs: feature propagation is conducted under different dynamics and is eventually combined to produce the finally prediction. The core idea of these methods is based on the hypothesis that rather than relying on a single  dynamic feature, which often leads to the OSQ problem, features obtained from multiple channels,  where similar types of data are grouped and propagated, can better preserve ``homogeneities'' and enable higher quality information flow to mitigate OSQ. For example, the MTGCN (multi-track graph convolution network) developed in \cite{pei2024multitrack} has the dynamic as 
\begin{align}\label{multi-tract}
    \mathbf H_T(\ell +1) = \widehat{\mathbf A} \mathbf H_T(\ell) + \alpha(\ell) \mathbf H_T(0), 
\end{align}
where $T \in \mathcal T(\mathbf Y)$ is the track of nodes that depends on their label (or pseudo-label) information \cite{pei2024multitrack}, and the final feature representation, e.g., for node $v$, is obtained by aggregating its information from the tracks that contain it with an additional MLP. It can be observed from Eq.~\eqref{multi-tract} that it prevents information mixing between nodes with different labeling information, and for nodes with the same information, the skip-connection term $\alpha\mathbf H_T(0)$ consistently reintroduces the feature variation into the system. Rather than identifying the responsible nodes based on their labels, \cite{new_choi2024panda} selects them according to their centralities. The core idea of \cite{new_choi2024panda} is that OSQ arises from the compression of  feature information, and one can mitigate OSQ by increasing the feature dimensionality of the responsible nodes (i.e., nodes with high centrality). Consequently, nodes with the same feature dimensionality are propagated together, separate from those with different dimensions, which is analogous to directing the information flow through distinct tracks.

}

\subsubsection{Other strategies}

In conclusion in the implicit rewiring section, we also spotlight recent works that resonate with the idea of reutilizing initial node features to alleviate the OSQ problem. For example, \cite{tortorella2022leave} applied a training-free reservoir computing paradigm to incorporate MPNNs with several additional propagations with the initial node feature after each layer without rewiring the initial graphs. Additionally,  rather than rewiring, the reweighting paradigm is considered in the work \cite{beaini2021directional}, in which a radius-restricted kernel is developed that assigns different edge weights to the subset of the nodes at a fixed distance radius (i.e., $r$). It is worth noting that different from the classic reweighting scheme, the method developed in \cite{beaini2021directional} further considers sign (directional) information that assigns the direction of the vector fields of each subset of the nodes. Through a well-crafted design of directional information, the flow of information is adjusted, thereby mitigating the OSQ problem. 

\revision{Furthermore, recent works \cite{zhao2025enahpool,finder2025improving} improve message-passing efficiency by introducing connections across graph clusters and in coarsened graphs, effectively transcending topological boundaries and enabling long-range interactions between distant nodes. Moreover, the OSQ problem has also been mitigated by assigning adjustments to the information flows; for example, \cite{shen2024handling} applies an element-wise maximizing operation to exploit all possible powers of the normalized adjacent matrix (with linear complexity) to construct a graph convolution operation. \cite{qiu2025gump} adjusts feature propagation by applying an unitary adjacency matrix, thereby preserving local connectivity. \cite{bause2024two} modifies and merges the neighboring tree structures of nodes to shift the feature dynamics while avoiding redundant computations, ensuring that only nodes within identical isomorphic subtrees can exchange information. 
}

{
\textit{Remark} (Jacobian-sensitivity on implicit rewiring).
Implicit rewiring can also be interpreted through Jacobian sensitivity. In a standard MPNN, the sensitivity of $h_i^{(\ell)}$ to $x_s$ is governed by powers of a sparse propagation operator, so distant nodes may have exponentially weak influence when few walks connect them. Implicit methods alleviate this by introducing additional long-range communication channels through attention, diffusion, spectral propagation, virtual nodes, or history-dependent mechanisms. Some models provide explicit sensitivity guarantees. For example, S$^2$GNN provides a positive lower bound on Jacobian sensitivity under suitable assumptions \cite{geisler2024spatio}, and S$^3$GNN further generalizes this result under relaxed conditions \cite{shi2026s3gnnefficientglobalmixing}. However, OSQ-specific Jacobian guarantees for general implicit rewiring methods remain limited, leaving a unified theory of implicit propagation, Jacobian sensitivity, and topological bottlenecks as an open problem. Overall, OSQ studies have shifted from ``fixed GNN, different views'' to ``different GNNs, fixed view'': early works analyze standard MPNNs via different OSQ-related quantities, while later works modify the graph or propagation mechanism to improve the long-range information flow.

\begin{tcolorbox}[breakable]

\textbf{Takeaway Message of this Section:} \textit
{{Implicit rewiring mitigates OSQ without explicitly changing the input graph. Instead, it creates denser or alternative communication channels through techniques such as diffusion, attention, multi-track propagation, or modified information transmission.}
}

\end{tcolorbox}

\revision{
\noindent \textbf{General Comparison Between Methods}\\
Spatial rewiring methods (e.g., curvature- or distance-based) primarily act on local neighborhoods, preserving community structure and nearby relations but often struggling with global bottlenecks unless many edges are added—risking graph densification and OSM. This locality–globality tension is well documented: spatial methods usually satisfy locality, while spectral methods trade some locality to improve global expansion/sparsity properties. Spectral rewiring targets global indicators (e.g., spectral gap), improving long-range connectivity but at higher conceptual and implementation cost. Implicit rewiring (diffusion/transformers/multi-track propagation) achieves similar effects via learned or dense propagation without explicit edge changes, though locality becomes less interpretable. From a computational perspective, curvature-based methods incur preprocessing costs,  whereas non-curvature methods vary between one-time precomputation and per-layer costs; we standardize these in Table~\ref{summary_table} and mark the stage explicitly. 
{Finally, it is worth noting that the connection between the methods reviewed in this section and the OSQ-related quantities in Section~\ref{preliminary} should be understood at the category level rather than as a one-to-one mapping between each method and a single quantity. A single rewiring or propagation mechanism may simultaneously affect adjacency powers, effective resistance, commute time, spectral expansion, and Jacobian sensitivity in a coupled and method-dependent manner; a complete characterization of these effects remains an open theoretical problem.}

}

\section{In Relation to OSM and Expressive Power}\label{sec:osq_theory}

\subsection{In Relation to Expressive Power}
Before we summarize the relationship between the OSQ problem and MPNNs' expressive power, we first note that although with slightly overlaps with the expressive power introduced in \cite{xu2018powerful}, the notion of the expressive power of MPNNs (or more generally GNNs) tends to be a general term. For example, the expressive power can be referred to as an MPNN's power of distinguishing non-isomorphic graphs \cite{xu2018powerful}, or the power of mixing node features \cite{new_di2024does}, etc. As we measure the OSQ problem via the sensitivity between node features at one specific layer and initial node feature (i.e., $\left \|\frac{\partial \mathbf{h}_i^{(r+1)}}{\partial \mathbf x_s}\right \|$), and $\mathbf{h}_i^{(r+1)}$ is generated from one specific MPNN model, and thus such sensitivity can in fact measure how powerful one MPNN model can mix the node features via a specific range of connection $r+1$. More specifically, if one MPNN is powerful enough, it shall have a larger $\left\|\frac{\partial \mathbf{h}_i^{(r+1)}}{\partial \mathbf x_s}\right\|$ so that all long-range relationship can be sufficiently captured to benefit the downstream regression/classification tasks, and vice versa. Motivated by this observation, \cite{new_di2024does} firstly linked the OSQ problem to the expressive power of MPNNs via both node and graph-level classification tasks. Specifically, the node-level expressive power can be defined as
\begin{align}\label{di_expressive_power}
    \mathrm{Mix}_{\mathbf Y}(i,s) \coloneqq \max_{\mathbf X} \max_{1 \leq \gamma, \delta \leq c_0} \left|\frac{\partial {\mathbf Y}_i(\mathbf X)[\gamma]}{\partial \mathbf x_s[\delta]}\right|,
\end{align}
in which we recall $\mathbf Y_i(\mathbf X)[\gamma]$ be the $\gamma$-th component of the node level map at node $i$ and $\mathbf x_s[\delta]$ the $\delta$-th entry of $\mathbf x_s$ at node $s$, respectively, and $\mathbf Y: \mathbb R^{N\times c_0} \rightarrow \mathbb R^{N\times c_0}$  the ground truth function that a MPNN is supposed to learn, \revision{and we denote node level expressive power as $\mathrm{Mix}_\mathbf Y$ to distinguish its graph level counterpart $\mathrm{mix}_{\mathbf Y_\mathcal G}$ in Eq.~\eqref{eq:glosq}. As we have illustrated for Definition~\ref{glosq}, both graph and node-level maximal mixing measure how expressive a model is in terms of mixing features with different nodes, i.e., \textit{expressive powers}.
}

It is worth highlighting that this expressive power is different compared to the OSQ score defined in Eq.~\eqref{osq_problem} which is the spectral norm of the Jacobian matrix between $\mathbf{h}_i$ and $\mathbf x_s$ pairs. 
Accordingly, \cite{new_di2024does} defined the OSQ via the 
\textbf{reciprocal} of $\mathrm{Mix}_\mathbf Y (i,s)$ and $\mathrm{mix_{\mathbf Y_\mathcal G}}(i,s)$, and the quantity of OSQ can be approximated by the quantity that depends on the MPNN's weight matrix, (weighted) graph adjacency matrix and a number of layer $\ell$, for more details, see definition E.2 in \cite{new_di2024does}\footnote{We note that we did not explicitly include the approximation equation due to the difference in terms of the definition of MPNNs in this paper compared to the one in \cite{new_di2024does}.}. In addition to the weights (in adjacency matrix) and depth ($\ell$), \cite{di2023over} also showed that both of them must be sufficiently large (according to the graph topology) to ensure mixing. For some tasks, the depth must exceed the highest commute time of the graph, and this aligns with the motivation of the recent work \cite{yu2023graph} and \cite{southern2025understanding}, in which a commute time-based MPNN is built for the node and graph-level classification problem.

\subsection{Trade-off between OSQ and OSM}
The trade-off between the OSM and OSQ problems is identified 
through the result of balancing the surgery applied on the graph to handle both problems. We recall the aforementioned observations for SDRF \cite{topping2021understanding} suggest that repeatedly adding edges to the graph will densify the graph and thus dilute the notion of community in which nodes have similar labels. That is, making the graph more localized and thus increasing the risk of the OSM problem, and this is the motivation of those spectral methods such as FOSR \cite{new_karhadkar2023fosr} which attempted to control the global connectivity measure (spectral gap) while rewiring. Another path to verify the trade-off between these two problems is through curvature sign, which has been shown that edges with highly positive curvature are responsible for the OSM problem whereas negatively curved edges are responsible for the OSQ problem from many recent studies such as \cite{nguyen2023revisiting,new_fesser2023mitigating}. \cite{shao2023unifying} unifies the OSQ and OSM problem through the \textit{spectral filtering} perspective and provides a provably correct solution on how one graph-based MPNN can prevent OSM problem while mitigating the OSQ problem. \revision{We refer readers to \cite{pei2024multitrack,new_wu2023difformer,maskey2023fractional,giraldo2022understanding} for detailed discussions of these trade-off, as the main focus of this paper is the OSQ problem.} In addition, we include more discussion on this promising aspect via the theoretical and empirical open questions in Section \ref{open_questions}.

\section{Empirical Strategies for  OSQ}\label{sec:empirical_strategy}
In this section, we summarize the current data benchmarks used for the recent literature to verify the effectiveness of OSQ mitigation methods. In addition, we also include a 
summary of the computation complexity of the existing methods.

\begin{table*}[t]
\begin{center}
\setlength{\tabcolsep}{7pt}
\renewcommand{\arraystretch}{1.5
}
\caption{Summary of the included spatial and spectral rewiring methods in terms of their indicators, computational complexity, baseline GNNs\& maximal number of layers and the accomplished tasks. We denote NC as node classification task, GC as graph classification task, LR as the long-term graph benchmarks and SCAL as the test of scalability.}\label{summary_table}
\vspace*{5pt}
\scalebox{0.75}{
\begin{tabular}{c @{\hspace{0.2cm}} | ll @{\hspace{0.2cm}} l |  @{\hspace{0.2cm}}  @{\hspace{0.2cm}}c @{\hspace{0.2cm}} c @{\hspace{0.2cm}} c @{\hspace{0.2cm}} c }
\toprule
\textbf{Methods} & \textbf{Indicator}  & \revision{\textbf{Complexity}$^\ddagger$\textbf{[Stage]}} & \revision{\textbf{Models($\#$layers)}}   &\multicolumn{4}{c}{\textbf{Tasks}} \\ 
&&&&\textbf{NC} &\textbf{GC} &\textbf{LR} & \textbf{SCAL} \\
\midrule
DIGL\cite{gasteiger2019diffusion}   &Degree Threshold                   &$\mathcal O(N)$ [Pre]            &GCN(4), GAT(4), GIN(4) &\ding{52}            & && \\
SDRF\cite{topping2021understanding}   &BFC                   &$\mathcal O(|\mathcal E|d^2_{\mathrm{max}})$ [Pre]            &GCN(2) &\ding{52}            & && \\
SJLR\cite{giraldo2022understanding}    &JLC                   &$\mathcal O(|\mathcal E|d_\mathrm{max})$ [Pre]          &SGC(4),GCN(4)             &\ding{52}            & &&\ding{52}\\
BORF\cite{nguyen2023revisiting}    &Ollivier                    &$\mathcal O(|\mathcal E| d^3_{\mathrm{max}})$ [Pre]       &GCN(9),GIN(9) &\ding{52}  &\ding{52} &\ding{52}       &\\
AFR3\cite{new_fesser2023mitigating}  &AFC                    &$\mathcal O(|\mathcal E|d_{\mathrm{max}})$ [Pre]                  &GCN(3),GIN(4)     &\ding{52}&\ding{52}&\ding{52}& \ding{52}     \\
GRAND\cite{chamberlain2021grand} &Diffusion weights      &$\mathcal O(|\mathcal E'|c_0)$ [Per-Layer] &GRAND(32)     &\ding{52} &&&\ding{52}\\
BLEND\cite{chamberlain2021beltrami}  &Positional encoding     &$\mathcal O(|\mathcal E|(d+d'^\dag))$ [Per-Layer] &BLEND(100$^*$)  &\ding{52}&&&\\
\textcolor{black}{$S^2$GNN \cite{geisler2024spatio}} &Virtual Nodes  & $\mathcal O(|\mathcal E|k)$ [Pre]$^\sharp$ & GCN(18)\&Spec(17)     &\ding{52} &\ding{52} &\ding{52} & \ding{52}\\

\hline
EGP\cite{deac2022expander}   &Cayley Graphs                   &$\mathcal O(\mathrm{log}(N))$ [Per-Layer]            &GIN(+5$^\delta$) &            &\ding{52} &&\ding{52} \\
GTR\cite{black2023understanding}     &Effective Resistance                    & $\mathcal{O}\!\left(N^{3}+k\,N^{2}\right)$ [Pre]
 &(R)-GCN(4), (R)-GIN(4)           & &\ding{52}&&\\
RLEF\cite{banerjee2022oversquashing}    &Cheeger Constant                    &$\mathcal O(N^2\sqrt{\mathrm{log}(N)}d^2_{\mathrm{max}})$ [Pre]     &GAT(6)            &\ding{52}$^{\dagger}$&&& \\
FOSR\cite{new_karhadkar2023fosr}    &Spectral gap                    &$\mathcal O(kN^2)$ [Pre]    &(R)-GCN(4), (R)-GIN(4)               &&\ding{52}&&\\
vDRew\cite{gutteridge2023drew}   &Delay mechanism                    &$\mathcal O(N|\mathcal E|)$ [Pre]    &GCN(23), GINE(23)        &\ding{52}&\ding{52}&\ding{52}& \\
LASER\cite{new_barbero2023locality}   &Connectivity measures                    &$\mathcal O(N^3)$ [Pre]  &R-GCN(5)     &\ding{52}&\ding{52} & \ding{52} &\ding{52}\\ 
\revision{GOKU \cite{liang2025mitigating}} & Spectrum & $\mathcal{O} (|\mathcal{E}'||\mathcal{E}| + |\mathcal{E}| {\rm polylog}(N))$ [Pre]   & GCN(4), GIN(4), GCNII(4) &\ding{52}&\ding{52} & &\ding{52}\\
\revision{MTMP \cite{pei2024multitrack}} & Multi-track & $\mathcal{O} (d|\mathcal T|N\eta)$ [Per-Layer] & GCN(64) &\ding{52}& & &\\
\hline 
\end{tabular}}
\end{center}
\vspace*{-5pt}
\footnotesize{\revision{$\ddagger$ Complexities are presented under the worst case scenario. For curvature-based methods, the complexity reports
their preprocessing cost. For non-curvature methods, we indicate the stage explicitly: [Pre] precomputation, [Per-layer] worst-case per forward/backward layer.
} $^\dag$ Additional run time complexity may induced from the number of function evaluations \cite{chamberlain2021beltrami}. 
$^*$ Maximum number of iteration steps \cite{chamberlain2021beltrami}. \revision{ $^\sharp$ The complexity is from pre-conducted eigendecomposition. }$^\delta$ GIN models with 5 additional EGP layers. 
$^{\dagger}$ neighbourhood matching problem. }
\end{table*}

\subsection{Synthetic Graphs}
\paragraph{Dumbbell and Ring-of-cliques Graphs for Neighbour Matching}
First introduced in \cite{alon2020bottleneck}, the dumbbell graph, often denoted as $K_n-K_n$, is a synthetic graph dataset comprising of two cliques $K_n$ joined by a bridge. The $d$-regular ring-of-cliques consists $m$ cliques each size $d+1$ connected in a ring after removing one edge from every clique
and connecting the endpoints of the removed edges \cite{mahlmann2005peer}. Both two graphs have low Cheeger constants and a high density of triangles, indicating that they are well-connected locally but poorly connected globally. In \cite{banerjee2022oversquashing}, a dumbbell graph with $N=50$ nodes and 4-regular ring-of-cliques with total $N=250$\footnote{with the label of the target nodes encoded as one hot random vector.} are leveraged for testing the rewiring dynamic of RLEF compared to the SDRF proposed in \cite{topping2021understanding}. We highlight that similar experiments with these two datasets have also been conducted via \cite{new_karhadkar2023fosr} 

\paragraph{RingTransfer}
The \texttt{RingTransfer} dataset is initially designed for testing the power of the method that is designed for enhancing the expressivity of GNN \cite{bodnar2021weisfeiler}. This dataset consists of chordless cycles, specifically rings, with a fixed size of $K$. In \cite{bodnar2021weisfeiler}, each graph within this dataset designates two nodes as source and target, both located at a predetermined distance from each other. The primary objective involves the target node producing a one-hot encoded label that corresponds to the source node. All other nodes within the ring are assigned a uniform constant feature vector. The challenge for a model is to effectively transfer the information originally associated with the source node to the opposite side of the ring where the target node is situated. In terms of the problem of mitigating OSQ, the dataset is further used in \cite{gutteridge2023drew} for testing the enhancing power of the proposed dynamic rewiring method with the so-called delay mechanism.

\paragraph{Erdos-Renyi random graph for scalability}
Different from the aforementioned synthetic graphs, the popular Erdos-Renyi random graph, which is generated from an artificial stochastic random model, is often utilized for testing the scalability of the rewiring algorithms \cite{new_barbero2023locality,new_karhadkar2023fosr,giraldo2022understanding}. For example, in \cite{new_barbero2023locality} the dataset is generated with $N = 10k$
nodes and Bernoulli probability $p = 10/N$, meaning that the graph has 10$N$ edges on average, and the computational time of LASER \cite{new_barbero2023locality} is reported.


\subsection{Classic Graph Benchmarks.}
For the real-world graph dataset, the classic citation networks (i.e., \texttt{Cora, Citeseer} and \texttt{Pubmed}) are the common choice for the node classification tasks. We highlight that for the methods targetting both the OSM and OSQ problems, both homophilic (connected nodes often with the same labels) and heterophilic graphs are selected, and the datasets can be found in \url{https://www.pyg.org/} with detailed summary statistics. In graph classification benchmarks, TUDataset, including \texttt{ENZYMES, MUTAG}, and \texttt{PROTEINS} \cite{morris2020tudataset}, is frequently utilized. In addition, several large-scale graph classification benchmarks such as ZINC \cite{sterling2015zinc} and \cite{hu2020open} are available for testing the scalability of the model. For the graph regression tasks, the dataset named \texttt{QM9} \cite{ramakrishnan2014quantum} which is a molecular multi-task dataset of over 130,000 graphs with on average 18 nodes each and a maximum graph diameter of 10, is usually preferred \cite{gutteridge2023drew}. We emphasize that the OSQ and OSM problems frequently manifest when the number of layers is large. It's noteworthy that there is considerable variation in the number of layers across studies \cite{topping2021understanding, giraldo2022understanding, nguyen2023revisiting, new_fesser2023mitigating}. This leads to one of the open questions of OSQ, i.e., how many layers are ideal for testing the OSQ problem and how to isolate the OSQ problem from OSM. 


\subsection{Long-range Benchmarks}
The OSQ problem assesses the capability of an MPNN to capture the long-range relationship between node pairs. One may prefer to verify those OSQ mitigation methods via the datasets that require long-range interactions. In \cite{dwivedi2022long}, five graph learning datasets namely \texttt{Pascal1VOC-SP}, \texttt{COCO-SP}, \texttt{PCMQ-Contact}, \texttt{Peptide-func} and \texttt{Peptides-struct}, which require long-range interaction for one MPNN to achieve strong performance, are presented. Specifically, \texttt{Pascal1VOC-SP} and \texttt{COCO-SP} are the datasets sourced from computer vision for node classification tasks. \texttt{PCMQ-Contact} is a dataset from quantum chemistry for link prediction tasks. Finally, \texttt{Peptide-func} and \texttt{Peptides-struct} are also the chemical data for the graph classification and graph regression tasks, respectively. In Table~\ref {summary_table} we list our selected papers that test their models on LRGB datasets.


\subsection{Underlying MPNNs/Baseline GNNs}
The rewiring methods aim to mitigate the OSQ problem along with the propagation of graph MPNNs. The choice of underlying MPNNs (known as baseline GNNs) is crucial for the practical implementation. We observe that the selection of this varies across the studies. For example, GCN \cite{new_kipf2017semi} for SDRF \cite{topping2021understanding}, SGC \cite{WuZhangSouza2019} and GCN for SJLR \cite{giraldo2022understanding} and R-GCN \cite{schlichtkrull2018modeling} for FOSR \cite{new_karhadkar2023fosr}, although ablation studies are conducted via some of included studies \cite{black2023understanding}. Given these baseline GNNs have different robustness on the OSM problem, their robustness on the OSQ problem varies as well due to the trade-off. Therefore a baseline GNN selection criteria is preferred to explore in future studies.

\subsection{Computational Complexity}
As an important aspect of the rewiring algorithm, the computational complexity is closely related to the model scalability. For example, curvature models usually compute the curvatures for all edges, and transformer-based models assign attention to all node pairs. In addition, the concern of complexity can also go to those dynamic rewiring methods as their computational cost tend to be high as the number of layer increases. Table \ref{summary_table} summarizes the reviewed spatial and spectral rewiring methods by their indicators, complexity, baseline GNN models together with the number of layers, and finally, the task conducted in their empirical studies.

\begin{tcolorbox}[breakable]
\textbf{Takeaway Message of this Section:} \textit{In general, datasets introduced for verifying OSQ mitigation methods are required to contain long-range dependency between nodes as provided in \cite{dwivedi2022long}. 
{This shall be accompanied by model-agnostic diagnosis, such as input-output Jacobian sensitivity. Empirical evaluation should also report complexity, scalability, and underlying MPNNs, since these choices affect whether OSQ mitigation is isolated.}
}
\end{tcolorbox}

\section{Open questions}\label{open_questions}


Building on earlier discussions, this section outlines open questions on OSQ, grouped into theoretical and empirical challenges, and presents potential directions for addressing them.

\subsection{Theoretical Open Questions}
\paragraph{OSQ Measurement}
Based on the current approaches, the OSQ score $\|\frac{\partial \mathbf{h}_i^{(r+1)}}{\partial \mathbf x_s}\|$ is only upper bounded as defined in Eq.~\eqref{osq_problem} and \eqref{osq_problem_2}. It is then unknown whether one MPNN can reach this bound after one enhancing method is applied. Accordingly, a lower bound\footnote{We assume here that MPNNs are sufficiently deep such that information can flow between the nodes at any distance.} of the Jacobian is desirable, \revision{while the only results in the current literature that we are aware of are the lower bounds obtained in \cite{new_di2024does} (for standard MPNN) and \cite{geisler2024spatio} (for $S^2$GNN) , where the former bound is strongly influenced by both the MPNN architecture (i.e., the minimum singular values of the weight matrix) and the graph topology (i.e., commute time), whereas the latter depends merely on Eq.~\eqref{s2gnn}. Furthermore, one can also observe that under the UniFilter (polynomial) layer proposed in \cite{huang2024universal}, the OSQ score is independent of the layer number $\ell$. Therefore, establishing a proper lower bound of the OSQ socre remains an critical open problem. 
}


\paragraph{Trade-off between OSQ and OSM}
As we have mentioned earlier, the discussion on the trade-off between OSQ and OSM is well established through those spatial indicators. In the theory of spatial methods, it is common to see that a very positive curved edge leads to the OSM problem while a negative curved edge is responsible for the OSQ problem. This phenomenon shows the fact that OSQ and OSM are entangled and may not be eliminated at the same time. Accordingly, one might align this phenomenon as the uncertainty principle utilized in quantum physics \cite{busch2007heisenberg} and wavelet analysis \cite{benedetto2021frame}. That is given two measures $\pi_\mathrm{OSM}$ and $\pi_\mathrm{OSQ}$, one expects that $\pi_\mathrm{OSM} \cdot \pi_\mathrm{OSQ} \geq \epsilon (\mathcal G, \phi, \psi, \ell)$, in which $\epsilon$ is a quantity that depends on the graph, and the MPNN elements.  

In addition, while their relationship is studied via spatial rewiring methods \cite{giraldo2023trade,nguyen2023revisiting,new_fesser2023mitigating}, there is a desire for an explanation via spectral methods. Interestingly, we observe the trade-off between spatial and spectral methods in terms of preserving graph locality can be in general treated as a mirror of reflecting the trade-off between the OSM and OSQ problems. In general, without considering the locality preservation, both two types of rewiring methods tend to significantly densify a graph, inducing the risk of the OSM problem but mitigating the OSQ problem. Therefore, preserving the locality will help the rewiring methods prevent the OSM problem. \revision{We highlight that recent work in \cite{jamadandi2024spectral} partially address this problem by showing that both OSM and OSQ problems can be addressed through a carefully designed edge deletion (rather than rewiring) scheme.}
To sufficiently quantify this phenomenon, especially through the changes of those spectral indicators (e.g., spectral gap in \cite{jamadandi2024spectral}), however, requires further mathematical verification in future studies.

\paragraph{How deep is deep?}
Recent analysis \cite{di2023over} indicates that the number of layers and the spectral norm of the weight matrix need to be sufficiently large and dependent on graph topology (such as commute time and effective resistance) to induce channel mixing.  However, increasing the number of layers of MPNNs leads to the OSM problem. Therefore, a method to find the optimized number of layers is preferable. We highlight that this target could be achieved by exploring both asymptotic \cite{digiovanni2023understanding} and non-asymptotic analysis \cite{new_wu2022non} of MPNNs. \revision{Furthermore, recall that in graph-level OSQ defined in Eq.~\eqref{eq:glosq}, one can see that if an MPNN requires fewer layers $\ell$ to reach the maximal mixing, i.e., $\mathrm{mix}_{\mathbf Y_\mathcal G}(i,s) =\max_{\mathbf x_i} \max_{1 \leq \gamma, \delta \leq c_0}\left|\frac{\partial^2 (\mathbf Y_\mathcal G(\mathbf X))}{\partial\mathbf x_i[\gamma] \partial \mathbf x_s[\delta]}\right| \,$, it will have a lower risk of the OSQ problem. Based on this observation, a lower bound of $\ell$ for quantifying the effect of virtual nodes is proposed in \cite{southern2025understanding} (with similar results in \cite{new_di2024does}) as follows. 
\begin{lem}\label{ell_number}
    Given $\mathcal G$ and node $i,j$ for which Eq.~\eqref{eq:commute-change} is negative, there are graph-functions with mixing between $i\neq s$ larger than some constants independent of $i,j$, such that an MPNN + virtual nodes of bounded weights to learn these functions, and the minimal number becomes 
    \begin{equation}
    \resizebox{0.49\textwidth}{!}{$
    \begin{aligned}
        \ell 
        \geq \frac{\mathrm{Com}_{\mathrm{vn}}(i,j)}{8}
        - \frac{|\mathcal E|}{8} 
        \sum_{i'=2}^N\frac{1}{\lambda_{i'}(\lambda_{i'}+1)}
        \left(\frac{N}{|\mathcal E|}\lambda_{i'} -1\right)
        (\mathbf u_{i'}(i)-\mathbf u_{i'}(j))^2
    \end{aligned}
    $}
    \end{equation}
\end{lem}
Lemma \ref{ell_number} suggests that when adding one virtual node that reduces the commute time in Eq.~\eqref{eq:commute-change}, the minimal number of layers required by MPNN + virtual node to learn a graph function to have maximal mixing is smaller than its original counterpart, showing an explicit quantification on the effect of virtual nodes enhanced MPNNs for avoiding GLOSQ problem. Nevertheless, the lower bound of $\ell$ for the node-level OSQ problem, and $\ell$'s relationship to the OSM problem, are still needed. 

}

\paragraph{How two types of rewiring methods affect each other?}
Although both spatial and spectral rewiring methods have shown remarkable enhancing power via mitigating the OSQ problem, how a spatial rewiring method affects the graph spectral quantities such as commute time, effective resistance as well and graph spectral gap is not yet fully understood. In fact, this realm has been rarely explored since  \cite{topping2021understanding} first explored how SDRF can affect the Cheeger constant of the graph \cite{giraldo2023trade}. On the other hand, one can observe that those spectral-based methods also lack interpretation on their rewiring effect on the graph spatial indicators, such as curvature, although preserving locality has gradually become a popular topic for spectral methods to explore \cite{new_barbero2023locality}. In fact, the concept of locality has not been formally defined quantitatively yet with its relationship with curvatures. Therefore, one promising further study direction is to explore the effect of spatial methods on graph spectral properties, and vice versa. We highlight that some recent works have obtained promising results between graph Ricci curvature and the eigen-distribution of the normalized Laplacian \cite{bauer2011ollivier}.

\paragraph{Equivalence between methods}
In this work, we summarized three possible strategies that can alleviate the OSQ issues. However, the relation between those methods is still unknown. We highlight that Di Giovanni et al. \cite{di2023over} partially resolve this challenge by investigating why spatial and spectral methods work through the perspective of effective resistance. However, from the perspective of the curvature, the exploration is not fully initialized. One inspirational aspect is to combine two metrics (i.e., effective resistance and curvature) together to build up a so-called effective resistance curvature \cite{devriendt2022discrete}, and explore whether this intermediate notion can bridge two types of methods.

\paragraph{Effects of mitigating OSQ}
Apart from OSQ and OSM, other GNN problems include poor performance on \revision{heterophilic graphs \cite{zheng2022graph,pei2024multitrack}, vanishing/exploding gradients \cite{rusch2022graph}, and under-reaching \cite{sun2022position}.} The relationship between OSQ, OSQ-mitigation strategies, and these issues remains under-explored. For example, whether locality-preserving rewiring is less effective than local rewiring on heterophilic graphs deserves investigation. Further, if rewiring is dynamic, i.e., providing a rewired graph at each MPNN layer, the resulting adjacency matrices can be viewed as a sequence of kernels. The question then becomes whether these kernels mitigate gradient vanishing/exploding in MPNNs. We note that this observation is similar to that in \cite{di2023over}.


\paragraph{A Unified Form of Definition and Bound}
Despite various definitions and bounds of OSQ have been developed, \cite{rusch2023survey} provides a unified formulation:

\begin{tcolorbox}[breakable]
\begin{defn}[OSQ]
    Let $\mathcal G$ be an undirected, connected graph with $\mathbf{H}^{(\ell)} \in \mathbb R^{N\times c_\ell}$ be the feature matrix generated from a graph-based MPNN after $\ell$ layers propagation. Let $\pi(\mathbf{h}^{(\ell)}_i, \mathbf x_s): \mathbb R^{c_\ell} \times \mathbb R^{c_0} \rightarrow \mathbb R_{\geq 0}$ be a sensitivity measure between node $i$ and $s$ with distance denoted as $d_\mathcal G(i,s) = r+1$ such that if $\ell < d_\mathcal G(i,s) $  then $\pi(\mathbf{h}^{(\ell)}_i, \mathbf x_s)$ = 0, then we define OSQ with its bound respect to $\pi$ as the layer-wise  decay of node pair sensitivity as $\pi(\mathbf{h}^{(\ell)}_i, \mathbf x_s) \leq C_1 (k) C_2(\widehat{\mathbf{A}}_{is},k)$.
    where $C_1$ is a certain valuedepending on the iteration time $k$ and $C_2$ on both connectivity of $i,s$ and $k$. 
\end{defn}
\end{tcolorbox}
For example, if $d_\mathcal G = r+1$ and the subgraph in the receptive field $\mathcal B$ is a binary tree, we have $C_2 =(\widehat{\mathbf{A}}^{r})_{is} =2^{-1} 3^{-r}$, suggesting a polynomial decay of the dependence of $i,s$ \cite{topping2021understanding}. In another example when $C_2 = \sum_{k=0}^{r+1} (\widehat{\mathbf{A}}^k)_{is}$, we recover the OSQ bound provided in \cite{black2023understanding}. Similarly, if $C_2 = \frac{d_\mathrm {max}}{2}\left(\frac{2}{d_\mathrm {min}} \left(r+2 - \frac{\mu^{r+2}}{1-\mu}\right)-R_{i,s}\right)$, we have the bound of OSQ via effective resistance proposed in \cite{black2023understanding}, and we recall that $R_{i,s}$ can be obtained through graph adjacency, i.e., Eq.~\eqref{rr_computation}. Although the alignment between definitions in terms of the definition of OSQ above has been observed, we expect more detailed mathematical discussion on this field from future studies, such as the forms of the sensitivity measures and the difference between the forms of $C_2$. 

\paragraph{Over-squashing in hypergraph neural networks}
Extending OSQ analysis to hypergraphs is an important future direction. Hypergraphs model higher-order interactions through hyperedges and have motivated architectures based on spectral convolution, clique-based approximations, unified message passing, and learnable multiset aggregation \cite{feng2019hypergraph,yadati2019hypergcn,huang2021unignn,chien2022you}. In hypergraphs, OSQ may arise not only from long node-to-node paths, but also from node--hyperedge--node aggregation, where many incident nodes are compressed into fixed-dimensional representations. Recent hypergraph benchmarks
show that advanced hypergraph models can sometimes be more susceptible to OSQ \cite{new_yadati2024oversquashing}. This suggests that graph-based indicators such as distance, curvature, and effective resistance, require nontrivial extensions to incidence structure, hyperedge size, overlap, and expansion schemes. Developing hypergraph OSQ measures and mitigation strategies remains underexplored.

\subsection{Empirical Open Questions}
\paragraph{Measure OSQ numerically}
Although existing works tackling OSQ verify the effectiveness of their approaches through various types of experiments (reviewed in Section \ref{sec:empirical_strategy}), there is no direct empirical measure for the degree of OSQ. It is necessary to directly quantify and compare the level of OSQ among different strategies. One natural measure is the Jacobian of trained models, which can be computed via auto-differentiation in a number of deep learning packages, like Pytorch. \revision{We refer readers to recent works in \cite{southern2025understanding,new_tori2025the} for further details.}

\paragraph{Isolating OSQ effect}

Similar to the open question on the trade-off between OSQ and OSM, empirically isolating the effect of either problem remains challenging, since both may occur simultaneously in deep MPNNs. A possible solution is to control the gradient flow of MPNN dynamics. Specifically, \cite{shao2023unifying,digiovanni2023understanding,han2022generalized} show that when MPNN dynamics are dominated by the largest eigenvalue of the graph Laplacian, i.e., high-frequency dominant (HFD), MPNNs sharpen node features from the beginning of propagation, thereby avoiding OSM. Moreover, Shao et al. \cite{shao2023unifying} and Han et al. \cite{han2022generalized} proposed controllable multi-scale MPNNs/GNNs to ensure HFD, while \cite{shao2023unifying} introduced a theoretically justified spectral filtering method to mitigate both OSQ and OSM under HFD. However, the manual adjustment in \cite{shao2023enhancing} leads to relatively large empirical variance, calling for more flexible control of model dynamics. The conditions in \cite{shao2023unifying} are necessary for mitigating both problems, whereas sufficient conditions are preferable in practice and should be further verified theoretically and empirically. \revision{Finally, thanks to LRGB \cite{dwivedi2022long}, the OSQ-mitigation effects of models can be compared on a relatively fair platform.}

\begin{tcolorbox}[breakable]
\textbf{Takeaway Message of this Section:} \textit{Measuring OSQ still represents the most fundamental challenge in OSQ research, especially when determining the lower bound of the OSQ score. 
{Future work should also clarify the OSQ-OSM trade-off and design empirical protocols that isolate OSQ effects across graph data and baseline MPNNs.}
}
\end{tcolorbox}

\section{Concluding Remarks}
In this work, we conducted a comprehensive review of the OSQ problem in MPNNs. In contrast to previous survey papers, we provided explicit mathematical formulations and bounds for OSQ, elucidated strategies for its mitigating, and delved into the current theoretical understanding of this problem. In addition, we offered a detailed description of the empirical strategies for validating the OSQ mitigation methods, including existing benchmark datasets, baseline graph-based MPNN models, discussion on computational complexity and scalability. We also emphasized various outstanding questions related to OSQ that remain unresolved to the best of our knowledge and suggested potential research directions. We hope our work paves the way for future research on the OSQ problem from both theoretical and practical perspectives.

\bibliographystyle{IEEEtran}
\bibliography{reference}
\vspace{-30pt}
\begin{IEEEbiography}[{\includegraphics[width=1in,height=1.25in,clip,keepaspectratio]{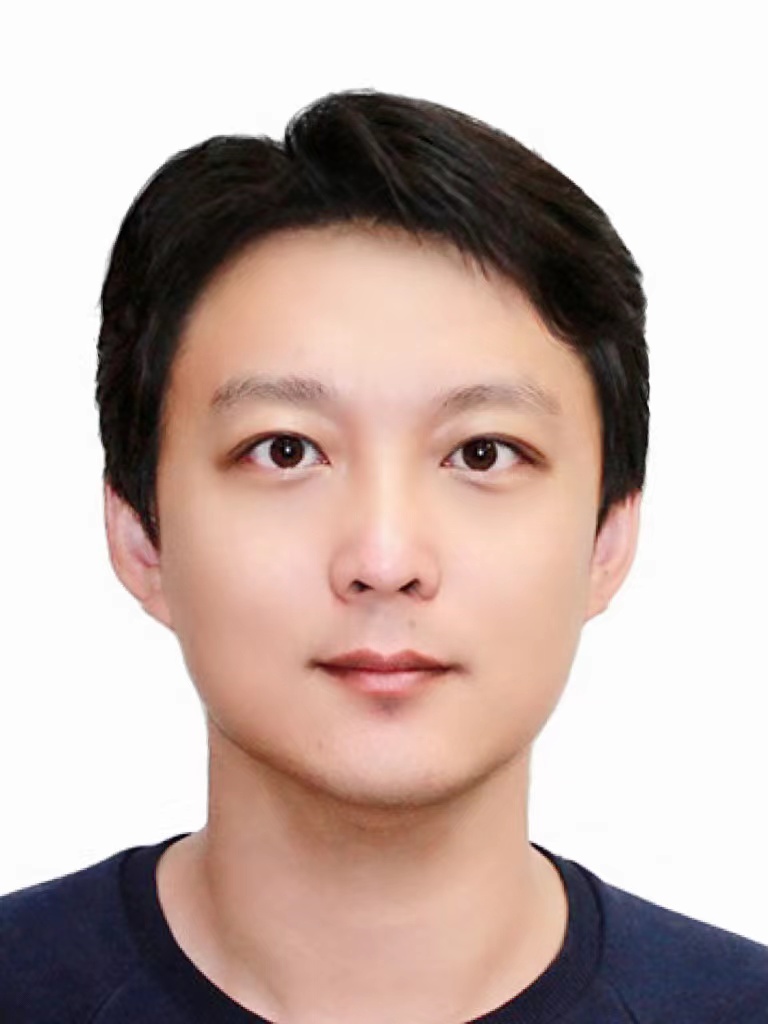}}]
{Dai Shi} is currently a postdoctoral research fellow at the University of Cambridge. He used to be one of the Chinese reserve team members for the international Mathematical Olympics. His current research interest involves AI4Science, generative models, stochastic (partial) differential equations and graph neural networks.
\end{IEEEbiography}

\vspace{-25pt}
\begin{IEEEbiography}[{\includegraphics[width=1in,height=1.25in,clip,keepaspectratio]{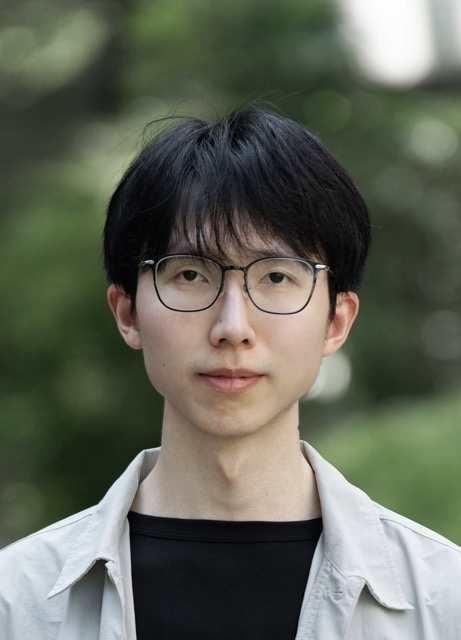}}]
{Andi Han} received the BCom. (Hons.) in business analytics from the University of Sydney in 2019. He received his Ph.D. degree from the University of Sydney in 2023. He is currently Lecturer at University of Sydney, School of Mathematics and Statistics. His current research interest involves generative models, efficient machine learning, and geometric machine learning.
\end{IEEEbiography}

\vspace{-30pt}
\begin{IEEEbiography}[{\includegraphics[width=1in,height=1.25in,clip,keepaspectratio]{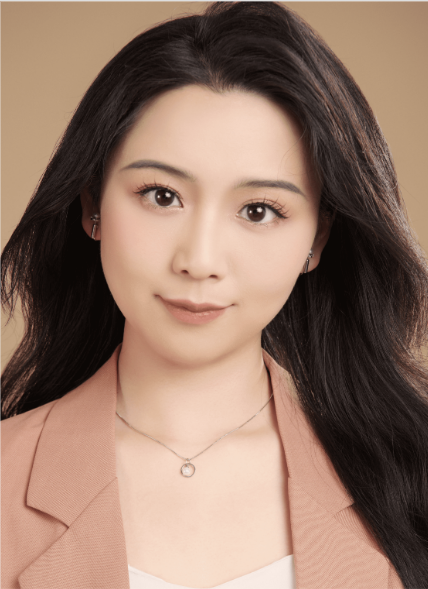}}]
{Lequan Lin} received the BCom. (Hons.) in Business Analytics at the University of Sydney Business School in 2022. She is currently a Ph.D. student in the Discipline of Business Analytics at the University of Sydney Business School. Her recent research interest involves large language models, graph neural networks and generative models.
\end{IEEEbiography}

\vspace{-30pt}
\begin{IEEEbiography}[{\includegraphics[width=1in,height=1.25in,clip,keepaspectratio]{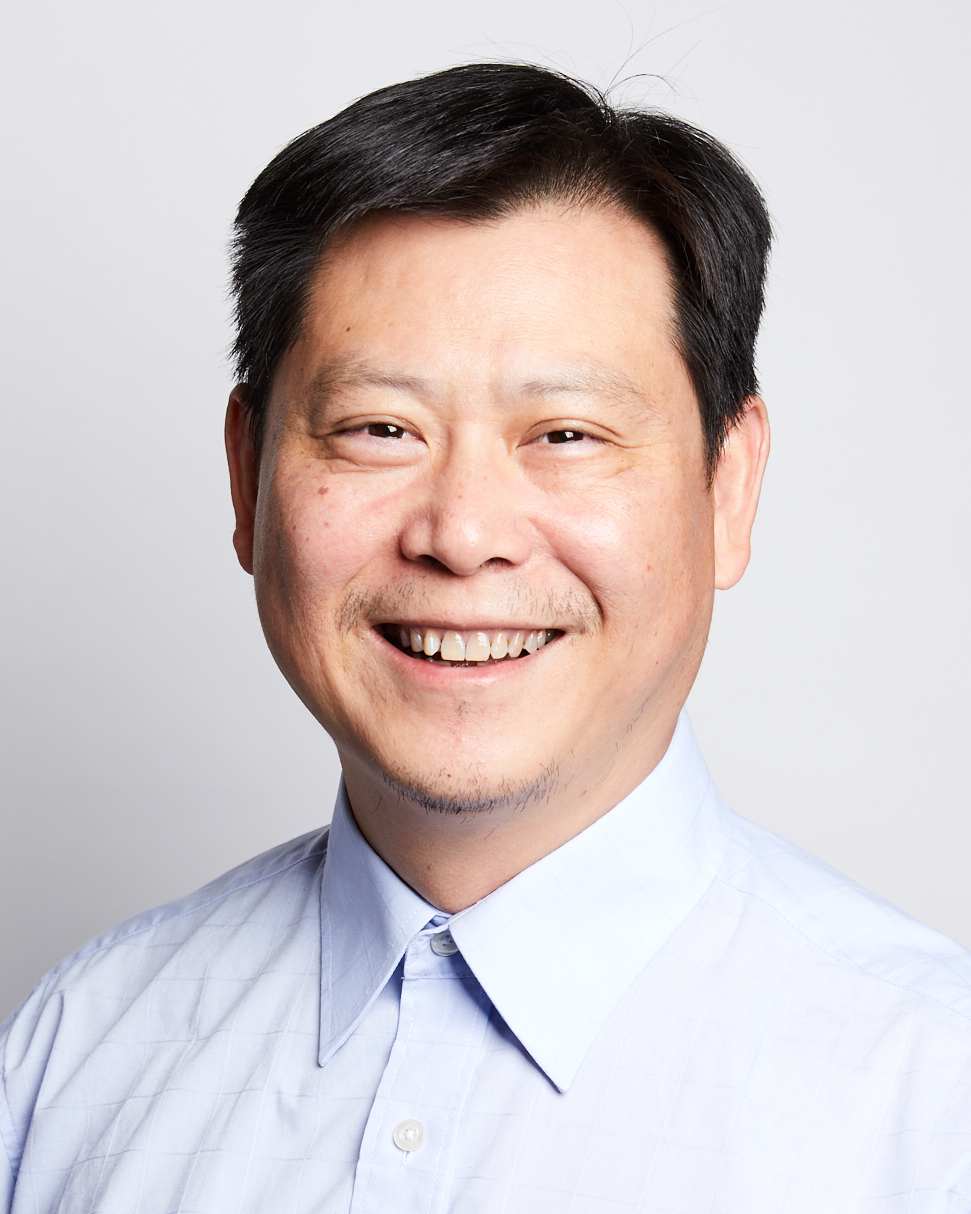}}]
{Yi Guo} received a B. Eng. (Hons.) in electrical engineering from the North China University of Technology in 1998, and an M. Eng. in automatic control from Central South University in 2002. From 2005, he studied Computer Science at the University of New England, Armidale, Australia, focusing on dimensionality reduction for structured data with non-vectorial representation. He received a Ph.D. degree in 2008. Between 2008 and 2016, he was with CSIRO, working as a computational statistician on various projects in spectroscopy, remote sensing, and materials science. He joined Western Sydney University in 2016. His recent research interests include machine learning, computational statistics, and optimization.
\end{IEEEbiography}

\vspace{-30pt}
\begin{IEEEbiography}[{\includegraphics[width=1in,height=1.25in,clip,keepaspectratio]{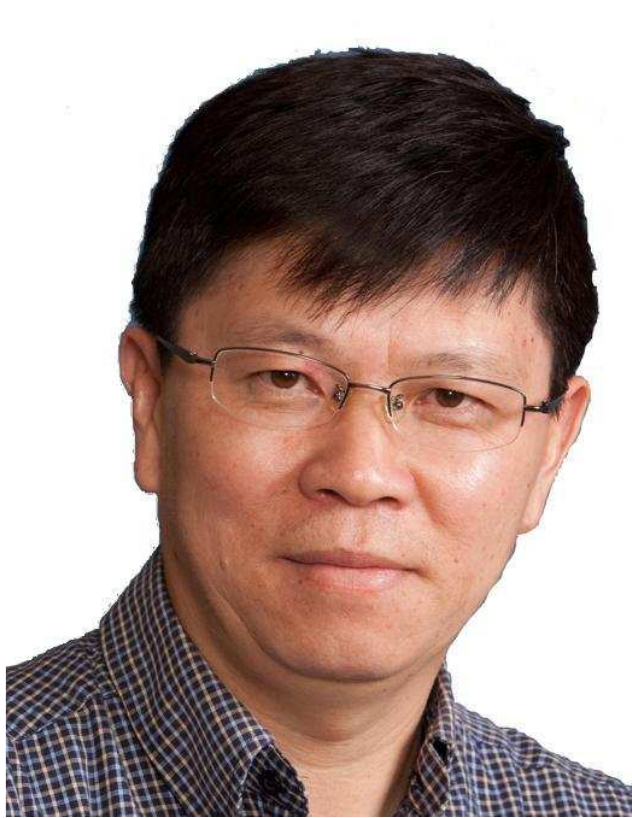}}]{Junbin Gao} graduated from Huazhong University of Science and Technology (HUST), China in 1982 with a BSc in Computational Mathematics and obtained his PhD from Dalian University of Technology, China in 1991. He is  Professor of Big Data Analytics in the University of Sydney Business School at the University of Sydney and was a Professor in Computer Science in the School of Computing and Mathematics at Charles Sturt University, Australia. He was a senior lecturer, a lecturer in Computer Science from 2001 to 2005 at the University of New England, Australia. From 1982 to 2001 he was an associate lecturer, lecturer, associate professor, and professor in the Department of Mathematics at HUST. His main research interests include machine learning, data analytics, Bayesian learning and inference, and image analysis.
\end{IEEEbiography}

\vfill

\end{document}